\documentclass{article}
\PassOptionsToPackage{numbers, compress}{natbib}

\usepackage{multirow}
\usepackage[preprint]{neurips_2026}

\usepackage[utf8]{inputenc} 
\usepackage[T1]{fontenc}    
\usepackage{hyperref}  
\hypersetup{hidelinks}
\usepackage{url}            
\usepackage{booktabs}       
\usepackage{subcaption}
\usepackage{caption}
\usepackage{amsfonts}       
\usepackage{nicefrac}       
\usepackage{microtype}      
\usepackage{xcolor}         
\usepackage{wrapfig}
\usepackage{float}

\usepackage{amsmath,amsfonts,bm}

\def\eqref#1{equation~\ref{#1}}

\def\1{\bm{1}}

\DeclareMathAlphabet{\mathsfit}{\encodingdefault}{\sfdefault}{m}{sl}
\SetMathAlphabet{\mathsfit}{bold}{\encodingdefault}{\sfdefault}{bx}{n}

\usepackage{url}
\usepackage{pifont}
\usepackage{graphicx}
\usepackage{subcaption}
\usepackage{tabularx}

\newcommand{\mcbheader}{%
\begingroup
\setlength{\tabcolsep}{0pt}
\footnotesize
\begin{tabular}{*{9}{>{\centering\arraybackslash}p{0.1111\textwidth}}}
&
\multicolumn{4}{c}{Prediction} &
\multicolumn{4}{c}{Absolute error} \\[0pt]
\cmidrule(lr){2-5}
\cmidrule(lr){6-9}
\shortstack{Ground\\truth} &
GAIA & LNO & NGF & CORAL &
GAIA & LNO & NGF & CORAL
\end{tabular}
\endgroup
}
\usepackage{enumitem}

\setlist[itemize]{leftmargin=*}
\setlist[enumerate]{leftmargin=*}
\title{GAIA: Geometry-Adaptive Operator Learning for Forward and Inverse Problems}

\author{
Meenakshi Krishnan \\
University of Maryland, College Park \\
\texttt{mkrishn9@umd.edu}
\And
Pranav Pulijala\thanks{Significant technical contribution} \\
University of Maryland, College Park
\And
Ke Chen \\
University of Delaware
\AND
Haizhao Yang \\
University of Maryland, College Park
\And
Ramani Duraiswami \\
University of Maryland, College Park
}

\begin{document}

\maketitle

\begin{abstract}
Operator learning for partial differential equations (PDEs) on arbitrary geometries builds fast neural surrogates for large-scale simulation. 
While recent geometry-adaptive neural operators have made substantial progress, they are mainly designed for forward problems in which inputs and outputs share the same spatial domain. This limits their applicability for boundary value problems (BVPs) and inverse problems, where inputs and outputs may live on different domains.
We introduce the Geometry-Adaptive Integral Autoencoder (GAIA), an operator learning model that encodes the domain boundary and the interior field distribution into geometry tokens, and conditions integral transform layers on these tokens via cross-attention, allowing the kernel to adapt locally to geometric features. This yields an architecture capable of learning solution operators for forward and inverse problems on arbitrary domains, solving each in a single pass at inference without per-domain optimization, iterative refinement, or graph construction.
We evaluate GAIA on seven 2D and 3D benchmarks, four of which are new or substantially extended benchmarks for inverse problems and BVP: electrical impedance tomography, optical tomography, 3D Darcy flow on varying geometries, and a modified setting of the Poisson BVP on mechanical components benchmark (MCB). GAIA sets new state-of-the-art results on every inverse and BVP task, reducing median relative $L^2$ error by 64\% on airfoil flow reconstruction and 22\% on EIT relative to the next best amortized method, and outperforming all baselines on every shape category of MCB. On other forward problems, GAIA is competitive with specialized solvers while maintaining stable accuracy across point resolutions on which transformer-based baselines degrade.

 \end{abstract}
\vspace*{-12pt}
\section{Introduction}
\vspace*{-3pt}
Partial differential equations are used to model problems across science and engineering, from fluid dynamics \citep{temam2024navier,arun2022unified} to medical imaging \citep{borcea2002electrical}. Traditional solvers are accurate but expensive, particularly in multi-query settings such as inverse problems, optimal design, and uncertainty quantification, where the forward model is invoked thousands of times. Early neural methods developed to address these costs, such as Physics-Informed Neural Networks (PINNs) \citep{raissi2019physics} and deep inverse priors \cite{dittmer2020regularization}, are instance-specific, requiring retraining for every new boundary condition or parameter change.

To overcome the need for repeated retraining, operator learning models instead learn maps between function spaces directly. However, classical neural operators such as FNO \citep{li2020fourier} and DeepONet \cite{lu2019deeponet} require regular grids or fixed discretizations. Many problems in structural analysis, medical tomography, and aerodynamic design are posed on domains whose geometry varies across problem instances, in both forward and inverse settings. A broadly applicable neural proxy must thus apply to arbitrary varying geometries without retraining.



Recent geometry-adaptive operators, such as GAOT~\citep{wen2025geometry}, GINO~\citep{li2023geometry}, Transolver~\citep{wu2024transolver}, handle unstructured point clouds via graph neural networks or transformer attention. However, these methods are mostly designed for forward modeling. Some, such as Transolver, require the input and output to share the same spatial discretization, precluding problems where the input and output live on different point sets - as in BVPs and inverse problems with boundary measurements. Applying these architectures to the latter requires computationally expensive iterative optimization loops over forward solvers. Others, such 
as GAOT, are designed and evaluated exclusively for forward problems. While GINO can in principle  handle inverse problems, its reliance on graph message passing incurs substantial memory and computational overhead.


We introduce the Geometry-Adaptive Integral Autoencoder (GAIA), a unified architecture for solving both forward and inverse problems on arbitrary domains. Building on IAE-Net~\cite{ong2022integral}, we introduce a dual-pathway tokenization that encodes both the domain boundary and the interior field distribution, and condition the integral kernels on these tokens via multi-head cross-attention. Unlike methods that encode geometry implicitly through per-sample neural field fitting~\cite{serrano2023operator} or require expensive graph construction~\cite{li2023geometry, wen2025geometry}, this provides explicit, spatially adaptive geometric context at each integration point. The query structure accommodates problems where the input and output live on different spatial domains, enabling efficient single-pass solution of inverse problems without iterative optimization.

In summary:
\begin{itemize} \item \textbf{Unified forward/inverse geometry-adaptive framework.} We propose GAIA, a geometry-adaptive integral operator that conditions learned kernels on domain geometry via cross-attention, enabling both forward and inverse problems on arbitrary geometries to be solved in a single pass.

\item \textbf{New benchmarks for varying-geometry problems.}  We introduce four new or substantially extended benchmarks for varying-geometry problems, including inverse problems such as Electrical Impedance Tomography (EIT) and Optical Tomography (OT).
All datasets are publicly available.
\item \textbf{Empirical validation.} We validate GAIA on seven time-independent PDE benchmarks spanning 2D and 3D problems. GAIA achieves state-of-the-art on all inverse and BVP tasks, reducing median relative $L^2$ error by 22\% on EIT and 64\% on airfoil relative to the next best method, while remaining competitive on forward benchmarks with stable accuracy across resolutions where transformer baselines degrade.
\end{itemize}

\section{Related Work}
\vspace{-3pt}
\subsection{Operator Learning}
\vspace{-2pt}
Early scientific machine learning methods such as Physics-Informed Neural Networks~\cite{raissi2019physics, karniadakis2021physics, yu2018deep, sirignano2018dgm, yu2022gradient, wang2021understanding} and CNN-based surrogates~\cite{zhu2018bayesian, bhatnagar2019prediction, adler2017solving} are either instance-specific or restricted to regular grids; we focus on operator learning. Neural operators learn mappings between function spaces, and can, for certain classes of PDEs, do so efficiently without suffering from the curse of dimensionality \cite{chen2025error}. DeepONet~\cite{lu2019deeponet} uses a branch-and-trunk decomposition but requires inputs sampled on a fixed grid. FNO~\cite{li2020fourier} parameterizes integral kernels in the spectral domain via the Fast Fourier Transform (FFT), achieving good performance on regular grids. Numerous extensions improve scalability and discretization handling through low-rank or factorized kernels~\cite{kovachki2023neural, tran2021factorized, rahman2022u, kossaifi2023multi}, graph message passing~\cite{li2020neural, li2020multipole}, and wavelet bases~\cite{tripura2022wavelet}. IAE-Net \cite{ong2022integral} achieves discretization invariance through a stack of integral autoencoder blocks with data-driven kernels, but its kernels depend only on local coordinates and function values with no access to global domain geometry.


For inverse problems, neural operators have been used either as differentiable surrogates within iterative optimization, MCMC, or diffusion-based sampling loops~\cite{wu2022learning, wu2024uncertainty, cao2025derivative, kaltenbach2023semi, shysheya2024conditional}, or to directly learn amortized inverse maps from measurement-to-parameter data pairs, enabling single-pass inference~\cite{molinaro2023neural, long2024invertible}. The former does not avoid the cost of iterative sampling at inference. GAIA belongs in the latter group. Neural Inverse Operators~\cite{molinaro2023neural} learn operator-to-function mappings via compositions of DeepONet and FNO; invertible architectures such as iFNO~\cite{long2024invertible} jointly learn forward and inverse maps through shared invertible blocks. However, both are tied to fixed regular grids due to their reliance on FFT. Unlike these deterministic amortized inverse methods, which produce point estimates, Bayesian approaches characterize posterior
distributions over unknown parameters
\cite{stuart2010inverse,kaipio2005statistical}. Recent work has explored
probabilistic operator learning for inverse problems and
conditional generative models for PDE simulations
\cite{nelsen2025operator,baldassari2023conditional,yao2026guided}.



\begin{figure}
    \centering
  \includegraphics[width=\textwidth, trim={7.5cm 10cm  7.5cm 4.5cm},clip]{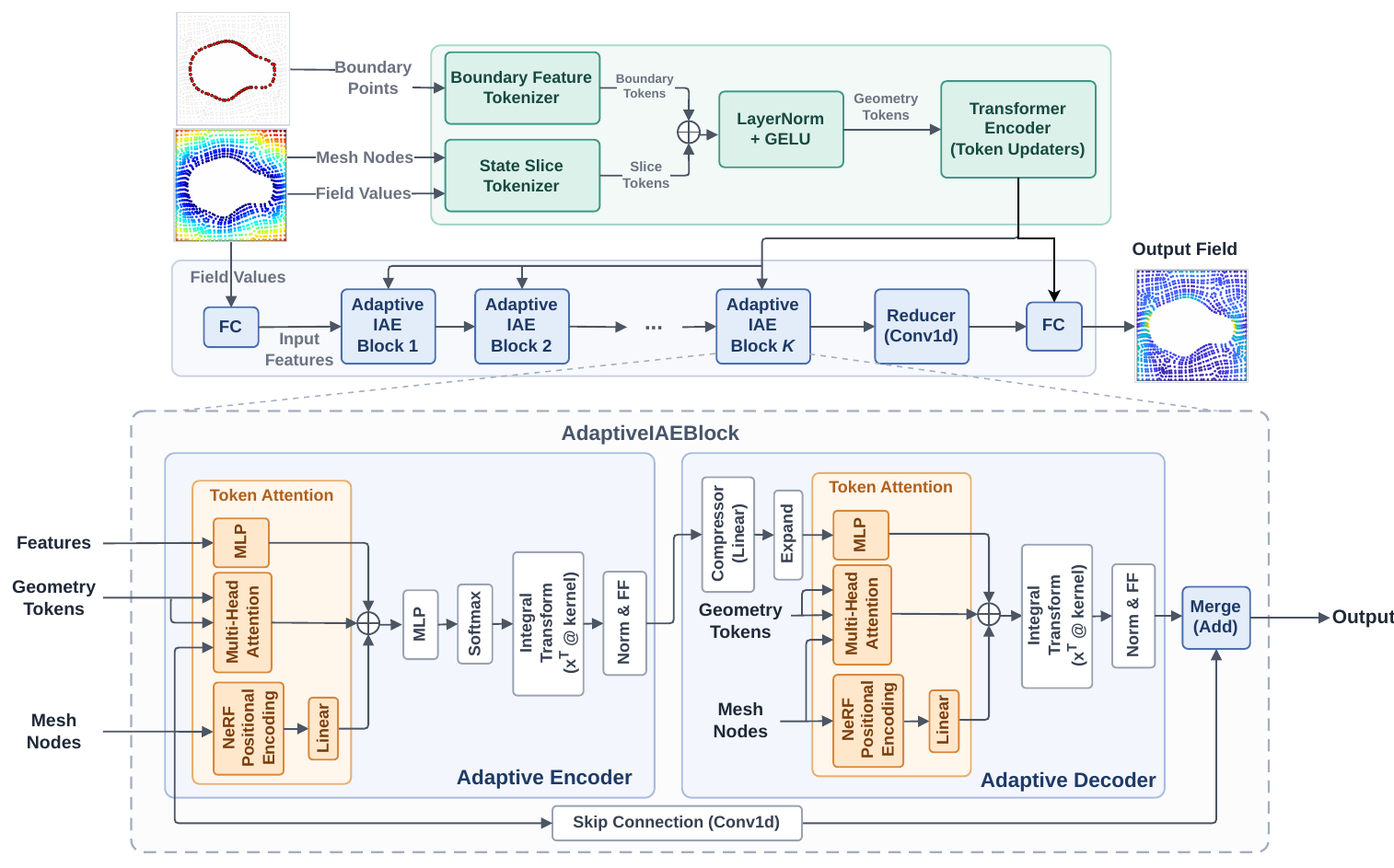}
   \caption{GAIA architecture. {\em Top:}\ Two complementary tokenizers produce geometry tokens from boundary points and interior mesh-field values, refined by transformer encoder layers. {\em Middle:}\ A stack of geometry-conditioned Adaptive IAE blocks with DenseNet-style skip connections maps input features to the output field. {\em Bottom:}\ Each block follows an encode–process–decode pattern with integral-transform kernels conditioned on geometry tokens via cross-attention. Separate encoder and decoder query sets enable inverse problems and BVPs.}
    \label{fig:collage}
\end{figure}

\vspace*{-4pt}
\subsection{Geometry-Adaptive Operators}
\vspace*{-4pt}
Geometry-adaptive operators take three broad approaches. \textbf{Coordinate-transform methods} like Geo-FNO~\cite{li2023fourier} learn a deformation from the physical domain to a regular latent grid; this works for topologically simple domains but fails on complex shapes. \textbf{Graph- and message-passing methods} including GINO~\cite{li2023geometry}, GAOT~\cite{wen2025geometry}, and RIGNO~\cite{mousavi2025rigno} handle unstructured point clouds natively but require per-sample graph construction that dominates inference cost. Other recent graph-based approaches include HAMLET \cite{bryutkin2024hamlet}, which employs graph transformers with a Galerkin cross-attention decoder, and AMG \cite{li2025harnessing}, which constructs multiscale and physics-informed graphs. MINO \cite{shi2025mesh} develops a mesh-informed operator backbone for functional generative modeling through flow matching. \textbf{Implicit-representation methods} \cite{catalani2024neural,du2024conditional} include CORAL~\cite{serrano2023operator}, which encodes fields as implicit neural representations and learns maps between latent codes. Transolver~\cite{wu2024transolver} and LNO~\cite{wang2024latent} are \textbf{attention-based methods} which project mesh points into a small set of learnable physical states via attention; broader transformer operators include GNOT~\cite{hao2023gnot} and OFormer~\cite{li2022transformer, xiao2023improved}. There are also specialized architectures such as Neural Green's Function \citep{yoo2025neural} for symmetric linear PDEs.


Among these, only LNO has been demonstrated on inverse tasks with differing input-output domains; CORAL and GeoFNO demonstrate iterative optimization over their forward surrogates for inversion. GAIA differs from all three: unlike CORAL it does not require per-sample neural-field fitting; unlike GINO it constructs no graph; and unlike LNO, which uses attention to compress spatial information into a fixed set of latent states, GAIA conditions the integral kernel itself on geometry at each query point via cross-attention. Our slice tokenizer adapts the soft-clustering aggregation of Transolver~\cite{wu2024transolver}; the boundary tokenizer and the use of both token pathways to condition integral kernels via cross-attention are novel, and the results demonstrate the architecture's efficacy.

\vspace*{-6pt}
\section{Methods}\label{sec:methods}
 \vspace*{-3pt}
We formalize operator-learning, recap the IAE-Net model~(\S3.1), and describe GAIA~(\S3.2).
\vspace*{-3pt}
\paragraph{Problem Formulation.}
We consider a generic time-independent PDE,
\begin{equation}
    \mathcal{D}(c, u) = q, \quad \forall x \in D \subset \mathbb{R}^d, 
    \quad \mathcal{B}(u) = u_b, \quad x \in \partial D,
\end{equation}
where $u : D \to \mathbb{R}^m$ is the solution, $c \in \mathcal{C}$ represents the physical 
parameters, $q \in \mathcal{F}$ is the forcing term,
$u_b$ prescribes the boundary values; and $\mathcal{D}$ and $\mathcal{B}$ 
are the differential and boundary operators, respectively. Also, let $\chi_D$ denote the domain indicator. Here $\mathcal{U}$, $\mathcal{C}$, and $\mathcal{F}$ denote suitable function spaces on $D$. Encapsulating 
the problem setup into $a = (c, q, u_b, \chi_D) \in \mathcal{A}$, the \emph{forward operator} $\mathcal{S} : \mathcal{A} \to \mathcal{U}$ maps $a \mapsto u$.

In the {inverse setting}, we do not have access to the full solution 
field $u$. We only have access to observables, often measured at 
the domain boundaries. This is formalized by a measurement 
operator $\mathcal{M} : \mathcal{U} \to \mathcal{V}$ mapping $u$ to an observation $v = \mathcal{M}(u)$. The generalized forward operator $\mathcal{F} = \mathcal{M} \circ \mathcal{S} : \mathcal{A} \to \mathcal{V}$ maps parameters directly to observables, and the \emph{inverse operator} $\mathcal{F}^\dagger : \mathcal{V} \to \mathcal{C}$ recovers $c = \mathcal{F}^{\dagger}(v, \chi_D)$ from these measurements.

Throughout, let $\Psi : \mathcal{X} \to \mathcal{Y}$ denote the target mapping ($\Psi \equiv \mathcal{S}$ in the forward case, $\Psi \equiv \mathcal{F}^\dagger$ in the inverse case). Our goal is to approximate $\Psi$ given access to data pairs $(f^{(i)}, g^{(i)})$ with $g^{(i)} = \Psi(f^{(i)})$.

\vspace*{-3pt}
\subsection{IAE-Net} 
 \vspace*{-3pt}
IAE-Net~\cite{ong2022integral} learns an operator $\Psi : \mathcal{X} \to \mathcal{Y}$ between function spaces on compact domains $\Omega_x \subset \mathbb{R}^{d_x}$ and $\Omega_y \subset \mathbb{R}^{d_y}$, from finite discretizations $S_x \subset \Omega_x$, $S_y \subset \Omega_y$ whose cardinalities may vary across samples. The architecture is a recursive stack of $L$ densely-connected autoencoder blocks, performing a series of transformations $f \to a_0 \to a_1 \to \dots \to a_L \to g$, with each intermediate function $a_i$ defined on $\Omega_a = [0,1]^{d_a}$. Each block performs an encode--process--decode transformation $a \to v \to u \to b$, where $v$ and $u$ live on a fixed latent domain $\Omega_z = [0,1]^d$, discretized on a grid $S_z = \{z_j\}_{j=1}^m$ independent of the input:
\begin{equation}
\small
v(z) = \int_{\Omega_x} \phi_1(a(x), x, z; \theta_1)\, a(x)\, dx,\ u(z) = \phi_0(v(z); \theta_0),\ b(y) = \int_{\Omega_z} \phi_2(u(z), y, z; \theta_2)\, u(z)\, dz.
\label{eqn:iae}
\end{equation}
Here, $\phi_1$ is the encoding kernel, $\phi_0$ a latent-space MLP, $\phi_2$ the decoding kernel, and $y \in \Omega_a$. The integrals are approximated as discrete sums, mapping variable-cardinality inputs to the fixed grid $S_z$. This allows IAE-Net to natively handle different input resolutions. Blocks are connected with DenseNet-style skip connections.

The kernels $\phi_1, \phi_2$ in equation \ref{eqn:iae} of IAE-Net depend only on local coordinates and the function values $a(x)$, with no access to global domain geometry. For elliptic PDEs, the Green's function governing the solution operator depends on the domain shape, so a geometry-agnostic kernel cannot capture this dependence. As a result, the same kernel is applied regardless of the domain geometry. Additionally, while IAE-Net's integral transform structure can in principle support differing input and output domains, this has not been explored for boundary-to-interior mappings.

 \vspace*{-3pt}
\subsection{Geometry-Adaptive Integral Autoencoder (GAIA)}
GAIA conditions IAE-Net's integral kernels on the domain geometry. We describe tokenizing the boundary points and interior fields, the conditioning mechanism, and the resulting blocks for forward and inverse problems. 
To address the geometry-agnostic limitation of IAE-Net described in \S3.1, the kernels are conditioned on a tokenized representation of the actual domain boundary and interior field distribution at inference time, with each spatial query point attending selectively via cross-attention, providing geometric context that is {spatially adaptive}. Decoupling query points from integration points lets the encoder integrate over $X_\text{bnd}$ while the decoder projects onto $X_\text{dom}$, enabling single-pass solve of inverse problems and BVPs.


\begin{figure}
    \centering
    \includegraphics[width=0.95\textwidth, trim={0cm 0cm  0cm 0cm},clip]{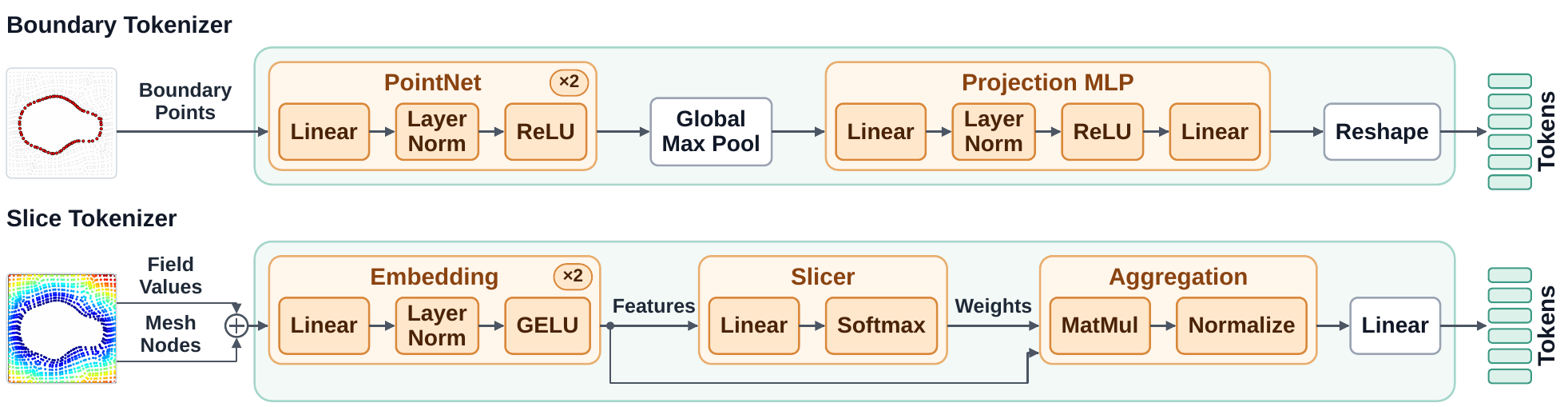}
    \vspace{1mm}
    \caption{{\bf The two tokenizer pathways:} GAIA encodes geometry through two complementary token sets. {\em Top -- Boundary tokenizer:} Boundary point coordinates are processed by a PointNet-style encoder with shared MLPs and global max-pooling to give permutation-invariant tokens that summarize the global domain shape. 
    {\em Bottom -- Slice tokenizer:} The interior mesh coordinates and their associated field values are jointly embedded and assigned to 
    soft clusters; mass-normalized aggregation produces tokens that summarize the spatially varying domain physics.} 
    \label{fig:narrow}
\end{figure}
\paragraph{Geometry Tokenization}
The geometry conditioning is constructed from two complementary token pathways: \emph{boundary feature tokens}, which encode the domain shape from boundary coordinates alone, and \emph{state slice tokens}, which encode the joint distribution of spatial coordinates and field values in the domain. Let $X_\text{bnd} \in \mathbb{R}^{M \times d}$ be the boundary/sensor coordinates and $X_\text{dom} \in \mathbb{R}^{N \times d}$ the interior mesh; let $U$ be the associated input field or measurements. The two pathways produce tokens that are concatenated into a unified $T \in \mathbb{R}^{K \times d_t}$; $K$ is the total token count, $d_t$ the token dimension.

\emph{Boundary feature tokenization.} The boundary coordinates $X_\text{bnd}$ are mapped to $K_\text{bnd}$ tokens $T_\text{bnd} \in \mathbb{R}^{K_\text{bnd} \times d_t}$ using a PointNet-style encoder~\cite{qi2017pointnet}, which applies shared MLPs followed by permutation-invariant global max-pooling, then projects and reshapes the resulting vector into $K_\text{bnd}$ tokens:
\begin{equation}
T_\text{bnd} = \text{Proj}\!\left(\max_{p \in X_\text{bnd}} \text{MLP}(p)\right).
\end{equation}



\emph{State slice tokenization.} Adapting the soft-clustering aggregation of Transolver~\cite{wu2024transolver} to inputs that include both spatial coordinates and physical values, we let $X \in \mathbb{R}^{P \times d}$ be a point cloud (instantiating as $X_\text{dom}$ for forward problems or $X_\text{bnd}$ for inverse problems) and $U$ the associated values. The concatenated state $[X, U]$ is mapped to $K_\text{slice}$ soft-cluster tokens via:
\begin{equation}
W = \text{Softmax}(\text{MLP}_\text{cluster}([X, U])), \quad H = \text{MLP}_\text{feat}([X, U]),
\end{equation}
\begin{equation}
T_\text{slice} = \text{diag}(W^\top \mathbf{1})^{-1} W^\top H,
\end{equation}
with mass-normalized aggregation; $\mathbf{1} \in \mathbb{R}^P$ is a vector of ones. A linear projection brings $T_\text{slice}$ to the standard token dimension $d_t$. The initial token set $T^{(0)} = [T_\text{bnd}, T_\text{slice}]$ is normalized and passed through GELU; between Adaptive IAE blocks, tokens are refined by standard transformer encoder layers, so that each block sees an updated token set $T^{(l)}$.

\paragraph{Geometry-conditioned integral transforms.}
GAIA modifies the encoding and decoding integral transforms (Eq. \ref{eqn:iae}) so that the learnable kernel is conditioned on the geometry tokens $T^{(l)}$. Spatial coordinates are first lifted via Fourier feature encodings $\gamma(x) \in \mathbb{R}^{2 d_F + d}$ with $F$ frequency bands~\cite{mildenhall2021nerf}. To inject geometric context into the kernel, we define a spatially-queried context vector $C(x)$ via cross-attention between the encoded coordinates and the geometry tokens:
\begin{equation}
C(x) = \text{Softmax}\!\left(\frac{(\gamma(x) W_Q)(T^{(l)} W_K)^\top}{\sqrt{d_k}}\right)(T^{(l)} W_V),
\end{equation}
where $W_Q, W_K, W_V$ are learnable projections and $d_k$ the scaling factor. The kernel is parameterized by an MLP taking the concatenated spatial features and geometric context:
\begin{equation}\label{eqn:cond}
K(x, z; \theta) = \text{MLP}_\text{kernel}\!\left([\gamma(x), \gamma(z), U(x), C(x)]\right),
\end{equation}
where $U(x)$ is the input feature at point $x$. The integrals in~(\ref{eqn:iae}) with this kernel are approximated as:
\begin{equation}
I(z) = \frac{1}{|X|}\sum_{x_j \in X} K(x_j, z; \theta)\, U(x_j),
\end{equation}
followed by a learned residual connection and GELU non-linearity.

\paragraph{Adaptive IAE Block.} GAIA uses these geometry-conditioned integral transforms as building blocks. Each Adaptive IAE block follows the encode--process--decode structure (\eqref{eqn:iae}) of IAE-Net, but with kernels conditioned on geometry tokens via cross-attention (Equation~\ref{eqn:cond}). The encoder maps from the $N$-point input to $m$ fixed latent modes, a pointwise MLP processes the latent representation, and the decoder maps back to $N$ points using the same conditioned kernel queried at the output coordinates. This bottleneck gives per-block complexity $O(Nm)$ with $m \ll N$.

\vspace*{-3pt}    
\paragraph{Main architecture.}
For forward problems where input and output share the same spatial domain, the Adaptive IAE blocks perform volume integrals over the domain mesh. For inverse problems and BVPs, an additional module first transfers information from boundary measurements to the interior.

\emph{Forward problems.} The input field $U \in \mathbb{R}^{N \times c}$ is defined on $X_\text{dom}$. The network applies $L$ Adaptive IAE blocks with dense skip connections: the input $H^{(l)}_\text{in}$ to the $l$-th block is the concatenation of all preceding outputs, compressed back to width $w$ via 1D convolution. The final prediction comes from passing the densely concatenated outputs through a reduction layer and a GELU MLP.

\emph{Inverse problems and BVPs.} Here the input is sparse measurements $U \in \mathbb{R}^{M \times c_\text{obs}}$ at boundary locations $X_\text{bnd}$, and the target lives on the dense interior mesh $X_\text{dom}$. To bridge this gap, an {Observation-to-Domain Decoder} first transfers information from the boundary to a fixed set of $K_\text{modes}$ learnable latent coordinates $Z \in \mathbb{R}^{K_\text{modes} \times d}$, then projects from the latent set onto the interior mesh. Concretely, the boundary data is embedded as $H_\text{bnd} = \text{MLP}_\text{in}(U) \in \mathbb{R}^{M \times w}$. A geometry-conditioned boundary integral lifts it to the latent set:
\begin{equation}
H_\text{mode}(z_k) = \frac{1}{|X|} \sum_{x_j \in X_\text{bnd}} K_\text{enc}(x_j, z_k; T^{(0)})\, H_\text{bnd}(x_j)\,  \qquad \forall z_k \in Z.
\end{equation}
A fixed-size MLP processes the latent representation, $\tilde{H}_\text{mode} = \text{MLP}(H_\text{mode})$, after which a second geometry-conditioned integral projects onto $X_\text{dom}$:
\begin{equation}
H^{(0)}(y_i) = \frac{1}{|Z|}\sum_{z_k \in Z} K_\text{dec}(y_i, z_k; T^{(1)})\, \tilde{H}_\text{mode}(z_k)\, \Delta z_k, \qquad \forall y_i \in X_\text{dom}.
\end{equation}
The resulting $H^{(0)} \in \mathbb{R}^{N \times w}$ enters the stack of Adaptive IAE blocks identically to the forward case.

 \vspace*{-6pt}
\section{Results}\label{sec:results}
 \vspace*{-6pt}

We evaluate GAIA against five geometry-adaptive baselines---GINO~\citep{li2023geometry}, CORAL~\citep{serrano2023operator}, Transolver~\citep{wu2024transolver}, GAOT~\citep{wen2025geometry}, and LNO~\citep{wang2024latent}---on seven benchmarks spanning forward, inverse, and BVP settings. Four of these benchmarks are new or extended contributions (Section~\ref{sec:benchmarks}); the remaining three are standard benchmarks. Transolver requires shared input/output discretization and is omitted from inverse and BVP comparisons. NIO~\citep{molinaro2023neural}, the closest amortized inverse-operator method, is FNO-based and operates on fixed regular grids; it cannot be applied to our varying-geometry benchmarks by construction. 
For the 3D Poisson BVP we additionally compare against NGF~\citep{yoo2025neural}, which is restricted to forward problems for linear symmetric PDEs.


For each task, we train GAIA using the prescribed architecture for forward (including BVPs) or inverse problems. All models are trained on a single NVIDIA A6000 GPU. We minimize the relative $L^2$ error,
\begin{equation}
\mathcal{L}(\theta) = \mathbb{E}_{f \sim \mu}\!\left[\frac{\|(\Psi_n(f; \theta) - \Psi(f)) \odot M\|_2}{\|\Psi(f) \odot M\|_2 + \epsilon}\right],
\label{eq:masked-loss}
\end{equation}
where $\odot$ denotes the Hadamard product. We use masking for varying mesh sizes; $M$ is a binary mask indicating valid interior nodes. Optimizers, learning rates, schedules, and full architectural hyperparameters for both GAIA and baselines are reported in Appendix~\ref{app:implementation_details}.

\vspace*{-6pt}
\subsection{Benchmarks}
\label{sec:benchmarks}
\paragraph{Inverse problems.}

\textit{Electrical Impedance Tomography (EIT)}. EIT
 is an imaging technique that reconstructs the internal conductivity $a({x})$ from boundary electrical measurements, solving an ill-posed \cite{calderon2006inverse} inverse problem governed by $\nabla \cdot (a({x}) \nabla u({x})) = 0$ on $\Omega \subseteq \mathbb{R}^2$. We apply $L{=}20$ Dirichlet excitation patterns at $M{=}272$ sensors on star-shaped domains and measure the Neumann response; the inverse operator maps the resulting Dirichlet-to-Neumann data $\Lambda_a$ to $a({x})$. We extend the fixed-geometry setting of~\cite{molinaro2023neural} to varying star-shaped domains.

\textit{Optical Tomography (OT)}.
OT is an imaging technique that recovers the scattering coefficient $\sigma_s({x})$ of a tissue-like medium from boundary light measurements. The transport of photons through the medium $\Omega \subset \mathbb{R}^2$ is governed by the stationary radiative transport equation (RTE):
\begin{equation}
    v \cdot \nabla_x u(x, v) + \sigma_t(x)\,u(x, v) = \sigma_s(x) \int_{S^{d-1}} \Phi(v \cdot v')\,u(x, v')\,dv', \quad x \in \Omega,
\end{equation}
where $u(x,v)$ is the photon density at position $x$ traveling in direction $v$, $\sigma_t = \sigma_s + \sigma_a$ is the total attenuation, and $\Phi$ is the scattering phase function. The boundary observable is the Albedo operator $\Lambda : L^1(\partial\Omega) \to L^1(\partial\Omega)$, which maps incoming illumination $\phi(x)$ to the outgoing flux $J_+(x) = \int_{v \cdot n_x > 0}(v \cdot n_x)\,u(x,v)\,dv$. The forward-peaked anisotropic scattering is modeled via the Henyey-Greenstein phase function. The Albedo operator is discretized as a dense source-to-receiver matrix. We use difference imaging~\cite{fan2019solving}: the Albedo operator $\Lambda_0$ of a homogeneous baseline ($\sigma_{s,0}{=}1$) is used as a reference, and the inverse operator maps the difference $\Delta\Lambda = \Lambda - \Lambda_0$ to the interior scattering perturbation $\delta\sigma_s({x})$. Domains are random convex pentagons. This is a new dataset.

\textit{Airfoil reconstruction.} We use the transonic airfoil dataset of~\cite{li2023fourier}, which contains steady-state solutions of the compressible Euler equations at $M_\infty=0.8$ over randomly deformed NACA-0012 profiles. We reformulate a sparse-to-full inverse reconstruction task: only 10\% of mesh nodes are observed, and at inference, the observations are corrupted by 1\% relative Gaussian noise. The model uses the sparse noisy observations to recover the full Mach field over the domain.


\begin{figure}
    \centering
    \includegraphics[width=\textwidth]{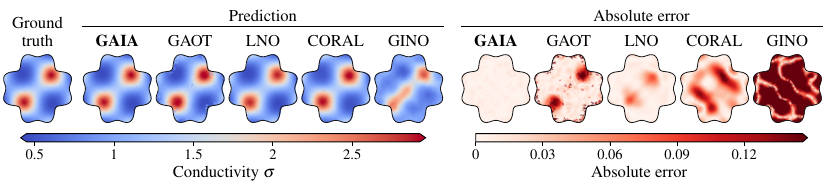}\\
    \includegraphics[width=\textwidth]{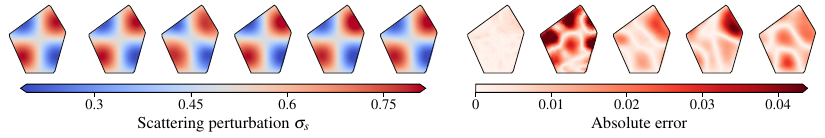}\\
    \includegraphics[width=\textwidth, trim=0 0cm 0 0mm, clip]{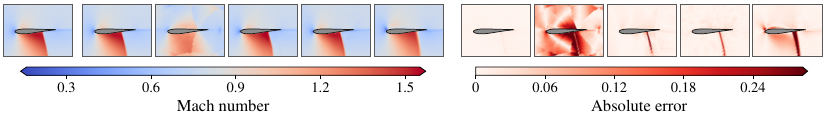}
    \caption{Model predictions on inverse benchmarks. \textbf{Top:} EIT conductivity reconstruction. \textbf{Middle:} OT scattering coefficient reconstruction. \textbf{Bottom:} Airfoil Mach field reconstruction from sparse noisy observations. GAIA captures fine-scale features like the conductivity patterns and transonic shock that baselines either mislocate (GINO) or diffuse (CORAL, LNO).  \vspace*{-6pt} }
    \label{fig:inverse_predictions}
\end{figure}

\vspace{-2mm}
\paragraph{Forward problems.}
\textit{3D Poisson BVP on mechanical components.}
We solve $-\Delta u = f$ on 3D mechanical parts from the MCB dataset~\cite{kim2020large}, spanning four shape categories: gears, nuts, fittings, and screws \& bolts. We adapt the setting and meshes of~\cite{yoo2025neural}, but formulate a pure BVP with fixed source and varying randomized Dirichlet boundary conditions, solved with FEniCS~\cite{logg2012automated}. The target operator thus maps the Dirichlet boundary function to the solution on the interior.

\textit{3D Darcy flow.}
We learn the map from a log-normal permeability field $\kappa( {x})$ to the pressure $u( {x})$ satisfying $-\nabla \cdot (\kappa \nabla u) = 1$ with homogeneous Dirichlet conditions on star-shaped 3D domains defined by random spherical harmonic expansions. 

\textit{Poisson-Gauss} and \textit{Elasticity}. These are standard  benchmarks with shared input--output discretizations from~\cite{mousavi2025rigno} and~\cite{li2023fourier} respectively. Details are in Appendix \ref{app:datasets}.

\vspace*{-6pt}
\subsection{Discussion}
\vspace*{-2pt}
\label{sec:inverse-results}

\paragraph{Inverse Problems} Table ~\ref{tab:inverse} reports median relative $L^2$ error on inverse and BVP benchmarks;  GAIA achieves the lowest error on every task. Against the next best amortized method, GAIA reduces error by 22\% on EIT (0.71\% vs.\ 0.91\% for LNO), 18\% on OT (1.45\% vs. 1.76\% for LNO), and 64\% on the Airfoil task (0.58\% vs.\ 1.59\% for CORAL). The Airfoil margin is the largest, likely because baselines struggle with the combination of noise, sparsity, and geometric variation. Visual reconstructions in the Appendix (Figures~\ref{fig:rte_predictions}-\ref{fig:pg_predictions}) show GAIA recovering fine-scale features (such as oscillatory conductivity patches in EIT, transonic shock structure in Airfoil) that baselines either smear or miss entirely. 

\begin{figure}[htb]
    \centering
    \mcbheader
    \vspace{-2mm}
    \begin{subfigure}{\textwidth}
        \centering
        \includegraphics[width=\textwidth, trim=0 0cm 0 4cm, clip]{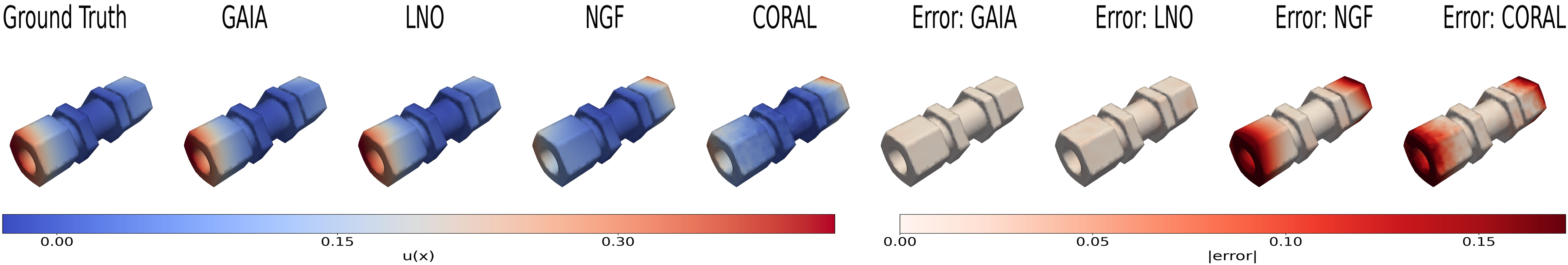}
        \label{fig:sub1}
    \end{subfigure}

  \vspace{-1mm}
    \begin{subfigure}{\textwidth}
        \centering
        \includegraphics[width=\textwidth, trim=0 0.3cm 0 6.1cm, clip]{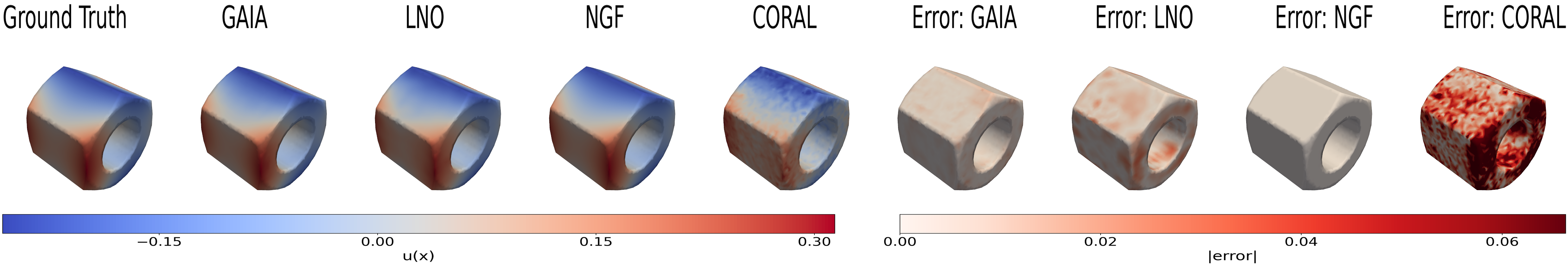}
        \label{fig:sub3}
    \end{subfigure}

 \vspace{-4mm}
    \begin{subfigure}{\textwidth}
        \centering
        \includegraphics[width=\textwidth, trim=0 0.3cm 0 6cm, clip]{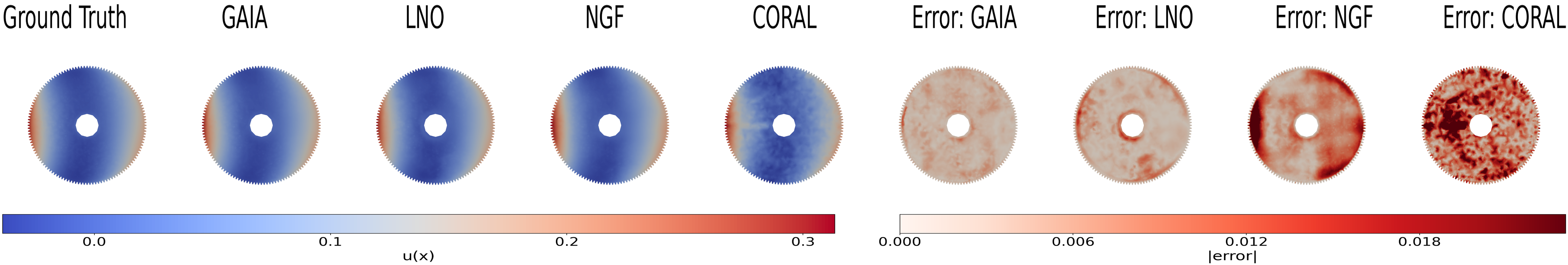}
        \label{fig:sub2}
    \end{subfigure}

    
    \caption{Results from the Mechanical Components Benchmark on the fitting, nut and gear categories. As shown in the error plots, GAIA reconstructs the solution with great fidelity across all shapes.}
    \label{fig:mcb}
\end{figure}

\paragraph{Boundary Value Problems} Figure~\ref{fig:mcb} visualizes representative reconstructions on some MCB shapes. On the 3D Poisson BVP (Table~\ref{tab:BVP}), GAIA outperforms all baselines on every shape category, with the largest gains on gears (3.63\% vs.\ 5.33\% for NGF) and screws \& bolts (3.45\% vs.\ 5.22\% for LNO). NGF is purpose-built for forward problems on linear symmetric PDEs; that GAIA exceeds it on every shape category, while remaining a general-purpose architecture, suggests the geometry-conditioned integral kernels carry meaningful advantage on real geometries.

\begin{table}[h]
\centering
\small

\begin{minipage}[t]{0.52\textwidth}
    \vspace{0pt}
    \centering
    \textbf{(a) Inverse Problems}\par\vspace{4pt}

    \begingroup
    \setlength{\tabcolsep}{3pt}
    \renewcommand{\arraystretch}{1.35}
    \begin{tabular}{lccccc}
        \toprule
        Dataset & GINO & CORAL & LNO & GAOT & GAIA \\
        \midrule
        EIT & 26.42 & 1.65 & 0.91 & 8.02 & \textbf{0.71} \\
        OT & 1.90 & 3.06 & 1.76 & 5.79 & \textbf{1.45} \\
        Airfoil & 4.79 & 1.59 & 2.49 & 9.53 & \textbf{0.58} \\
        \bottomrule
    \end{tabular}
    \endgroup
\end{minipage}
\hfill
\begin{minipage}[t]{0.46\textwidth}
    \vspace{0pt}
    \centering
    \textbf{(b) Mechanical Components BVP}\par\vspace{4pt}

    \begingroup
    \setlength{\tabcolsep}{3pt}
    \renewcommand{\arraystretch}{1.10}
    \begin{tabular}{lcccc}
    \toprule
    Shape & CORAL & NGF & LNO & GAIA \\
    \midrule
    Fitting & 18.81 & 11.25 & 14.88 & \textbf{10.30} \\
    Gear & 13.81 & 5.33 & 5.91 & \textbf{3.63} \\
    Nut & 20.69 & 5.34 & 6.95 & \textbf{4.95} \\
    Screws/Bolts & 20.47 & 6.31 & 5.22 & \textbf{3.45} \\
    \bottomrule
\end{tabular}
    \endgroup
\end{minipage}

\vspace{5pt}
\caption{Comparison on (a) inverse problems and
(b) the Mechanical Components BVP benchmark.
Median relative $L^2$ error (\%).
GAIA achieves the lowest error across all benchmarks.}
\label{tab:inverse_bvp}
\label{tab:inverse}
\label{tab:BVP}
\end{table}

\vspace{-10pt}
\paragraph{Other Forward Problems} Table~\ref{tab:forward} reports forward problem results on benchmarks where input and output share a discretization. GAIA achieves the lowest error on Poisson-Gauss (0.77\%) and is competitive with the leading method on Elasticity (1.39\% vs.\ 0.99\% for GAOT). On 3D Darcy flow, Transolver achieves lower error than GAIA at the training resolution (0.79\% vs.\ 1.24\%). This advantage, however, is contingent on evaluation matching the training discretization. As shown in \S\ref{sec:di}, both Transolver and GAOT degrade by an order of magnitude when evaluated at coarser resolutions on the Elasticity benchmark where they look strongest in-distribution; GAIA's accuracy is stable. We view this as an architectural trade-off: GAIA's integral-transform structure may prioritize discretization invariance over peak in-distribution accuracy on shared-mesh forward tasks. For forward problems where deployment-time mesh resolution is fixed and matches training, Transolver remains a good choice; for applications where resolution varies---which includes many multi-query settings---GAIA's robustness profile is preferable.
\vspace{-2mm}
\begin{table}[htbp]
    \centering
    \captionsetup{font=small}
    \caption{Comparison on forward benchmarks.
   Median relative $L^2$ error (\%). GAIA is competitive with baselines.}
    \label{tab:forward}
    \vspace{3pt}
    {
    \setlength{\tabcolsep}{6pt}
    \begin{tabular*}{\linewidth}{@{\extracolsep{\fill}}lccccc@{}}
        \toprule
        Dataset & GINO & CORAL & Transolver & GAOT & GAIA \\
        \midrule
        Poisson-Gauss & 1.21 & 4.36 & 1.22 & 1.37 & \textbf{0.77} \\
        Elasticity & 1.93 & 1.98 & 1.00 & \textbf{0.99} & 1.39 \\
        3D Darcy & 20.51 & 21.58 & \textbf{0.79} & 36.27 & 1.24 \\
        \bottomrule
    \end{tabular*}
    }
\end{table}

\vspace{-1mm}
\subsection{Discretization Invariance}
\label{sec:di}
\begin{wrapfigure}{r}{0.43\textwidth}
    \vspace*{-20pt}
    \centering
    \includegraphics[width=0.4\textwidth]{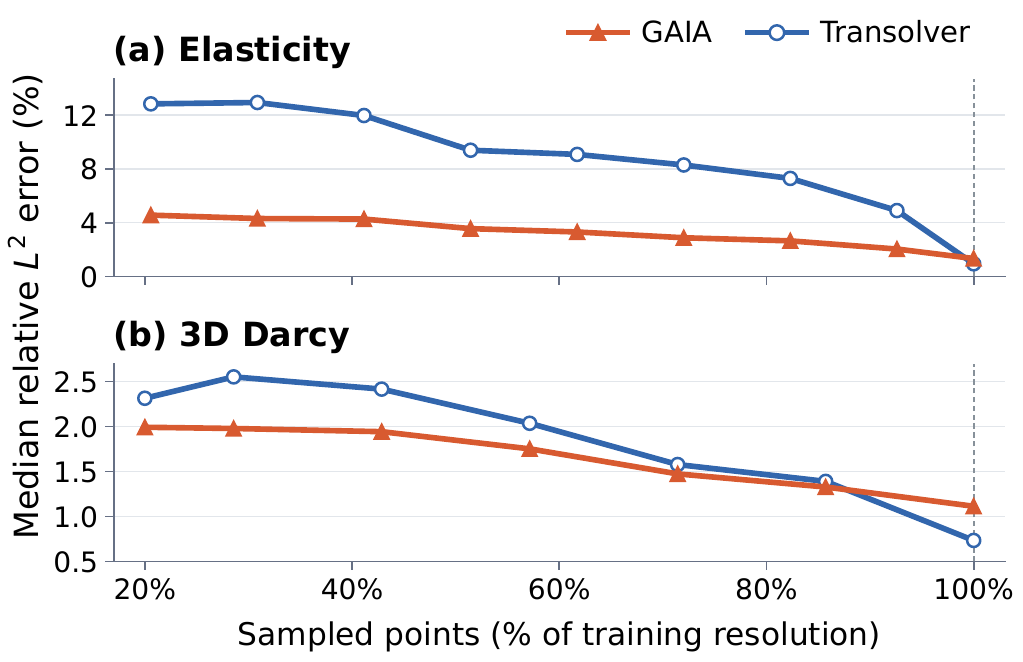}
    \captionsetup{font=small}
    \caption{Discretization invariance on Elasticity and 3D Darcy. GAIA maintains accuracy across resolutions while Transolver degrades. 
    }
    \label{fig:discretization_elasticity}
\end{wrapfigure}
A central claim of integral-transform operators like IAE-Net is \emph{discretization invariance} (DI): accuracy must be maintained when the model is evaluated at point-cloud resolutions different from training. We test this on Elasticity---the forward benchmark where Transolver and GAOT achieve better results than GAIA---by training each model at the full resolution and evaluating on coarser subsets obtained via Farthest Point Sampling. Figure~\ref{fig:discretization_elasticity} shows that GAIA maintains accuracy across all tested resolutions, while Transolver and GAOT (Appendix \ref{app:di}, Figure~\ref{fig:di_darcy}) degrade significantly away from the training resolution. The in-distribution Transolver-vs-GAIA gap on Elasticity reported in Table \ref{tab:forward} is an artifact of evaluating at the training resolution; under resolution mismatch, the comparison reverses. GAIA similarly outperforms Transolver at non-training resolutions on 3D Darcy.





\subsection{Efficiency analysis} 
Table~\ref{tab:timing_analysis} reports inference cost on EIT ($N=2000$ mesh nodes) on a single A6000 in FP32. Latency is the mean wall-clock time over 100 runs (batch size 1); throughput is measured at batch size 16. GAIA achieves the highest throughput among geometry-adaptive methods (365 samples/s) at competitive latency (16~ms) and modest peak memory (48~MB). The closest method on throughput is CORAL at 253 samples/s. Among graph-based methods, GINO incurs roughly $2\times$ GAIA's latency at $40\times$ the memory due to its dense FNO latent grid.  We additionally compare GAIA inference with classical FEM on the Mechanical Components Benchmark, highlighting the computational savings from avoiding per-instance mesh generation and numerical solves (Tab.~\ref{tab:fem_comparison}, Appendix~\ref{app:fem_comparison}).


\vspace{2mm}
\begin{table}[h!]
\centering
\begin{tabular}{lccccc}
\toprule
\textbf{Model} & \textbf{Params (M)} & \textbf{Latency (ms)} & \textbf{Memory (MB)} & \textbf{Throughput (samples/s)} \\
\midrule
GAIA & 3.02 & 15.78 & 47.60 & 364.70 \\
CORAL & 0.84 & 4.99 & 32.70 & 252.9 \\
GINO & 2.31 & 36.54 & 1886.50 & 27.3 \\
GAOT & 3.40 & 10.45 & 45.4 & 135.9\\
\bottomrule
\end{tabular}
\vspace{3mm}
\caption{Computational cost comparison on EIT.  GAIA has competitive latency and memory while maintaining the highest throughput among geometry-adaptive methods}
\label{tab:timing_analysis}
\end{table}

\begin{wraptable}{r}{0.45\textwidth}
\vspace{-5pt}
\centering
\caption{Scalability on 3D Darcy (single A6000 GPU).
Training measurements are per-step; inference uses
batch size 1.}
\label{tab:scaling_main}
 \small
\setlength{\tabcolsep}{3pt}
\begin{tabular}{rcccc}
\toprule
& \multicolumn{2}{c}{Training}
& \multicolumn{2}{c}{Inference} \\
\cmidrule(lr){2-3}\cmidrule(lr){4-5}
Points & ms & MB & ms & MB \\
\midrule
7K   & 57.8   & 930    & 14.18   & 102 \\
100K & 262.2  & 12,220 & 91.57   & 1,165 \\
400K & 1082.0 & 48,726 & 397.04  & 4,595 \\
1M   & OOM    & OOM    & 1002.37 & 11,449 \\
\bottomrule
\end{tabular}
\vspace{-5pt}
\end{wraptable}

We evaluate scalability on the 3D Darcy benchmark
(Table~\ref{tab:scaling_main}), observing approximately
linear growth in runtime and memory usage. GAIA supports
million-point inference on a single GPU, while training-step
memory limits the resolution to approximately 400K points.
A separately trained 100K-point model achieves 2.6\%
median relative $L^2$ error under a reduced training
budget. The single-GPU is the shared bottleneck documented in \cite{luo2025transolver++}, which addresses it via multi-GPU parallelism; the same standard techniques (gradient checkpointing, chunked query evaluation, parallelism) can extend GAIA's range. Full results and comparisons with Transolver
are provided in Appendix~\ref{app:scaling}.

\noindent

\newpage
\vspace*{-6pt}
\subsection{Noise Robustness}\vspace*{-6pt}\begin{wrapfigure}{R!}{0.4\textwidth}
    \vspace*{-18pt}\centering
    \includegraphics[width=0.38\textwidth]{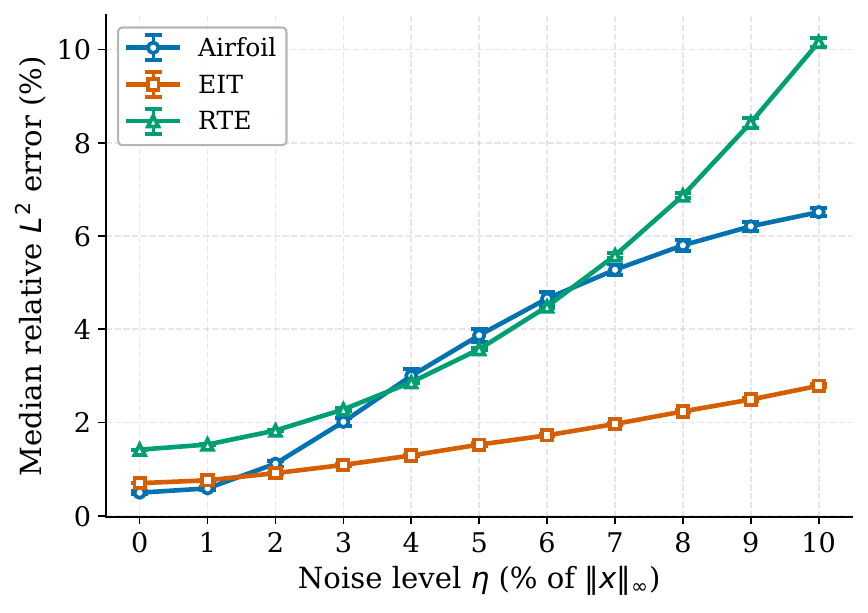}
    \caption{Graceful degradation in median relative $L^2$ error (\%) when increasing noise on inverse benchmarks.\label{fig:noise}}
     \vspace{-5pt} 
\end{wrapfigure}
We evaluate GAIA's robustness to test-time measurement noise by training once on clean data and evaluating at noise levels $\eta \in \{0\%, 1\%, 2\%, 5\%, 10\%\}$ using an additive Gaussian noise model scaled by the per-sample $L^\infty$ norm. Figure~\ref{fig:noise} shows graceful degradation across all three inverse benchmarks: at $\eta = 10\%$, error grows by roughly $13\times$ on Airfoil, $4\times$ on EIT, and $7\times$ on RTE relative to the clean baseline, with no sharp transitions or instability. The graceful-degradation behavior is consistent with what we would expect from an integral-transform architecture: noise in the input is averaged across the integration kernel rather than being amplified by point-wise operations.

\vspace*{-6pt}
\subsection{Ablations}
\vspace*{-6pt}
To evaluate the improvement over the IAE-Net backbone, we compare GAIA against IAE-Net at matched parameter count on Elasticity. GAIA achieves 1.39\% median relative L2 error, compared to 3.79\% for IAE-Net.
Table~\ref{tab:ablations} reports two ablations on Elasticity (forward) and Airfoil (inverse). \emph{Conditioning mechanism:} cross-attention strictly outperforms both concatenation and FiLM, with the largest gap on Airfoil (0.58\% vs.\ 5.91\% for concatenation, an order of magnitude). This confirms that spatially adaptive conditioning---each query point attending independently to the geometry tokens---is essential for capturing local geometric features; a single global summary vector is insufficient on tasks with complex spatially varying physics. \emph{Tokenizer pathways:} removing either the boundary tokens or the slice tokens degrades performance on both benchmarks. The two pathways encode complementary information: boundary tokens summarize global domain shape, while slice tokens capture spatially varying input physics. We further investigate architectural choices, including token refinement, boundary encoding, skip connections, positional encoding, and the observation-to-domain decoder, as well as hyperparameter sensitivity to token count and dimension (Tabs.~\ref{tab:architecture_ablation}-\ref{tab:ablation_num_tokens}, Appendix~\ref{app:ablate}). We find that no ablated variant consistently outperforms the full model and sensitivity to token hyperparameters is mild across the tested ranges.

\begin{table}[htbp]
    \centering
    \small
    \setlength{\tabcolsep}{4pt}

    \begin{minipage}[t]{0.48\linewidth}
        \centering
        \textbf{(a) Conditioning}\par\vspace{4pt}
        \begin{tabular*}{\linewidth}{@{\extracolsep{\fill}}lcc@{}}
            \toprule
            Method & Elasticity & Airfoil \\
            \midrule
            Concatenation & 1.77 & 5.91 \\
            FiLM & 1.88 & 0.88 \\
            Cross-attention & \textbf{1.39} & \textbf{0.58} \\
            \bottomrule
        \end{tabular*}
    \end{minipage}%
    \hfill
    \begin{minipage}[t]{0.48\linewidth}
        \centering
        \textbf{(b) Tokenizer pathways}\par\vspace{4pt}
        \begin{tabular*}{\linewidth}{@{\extracolsep{\fill}}lcc@{}}
            \toprule
            Variant & Elasticity & Airfoil \\
            \midrule
            No boundary & 2.28 & 1.06 \\
            No slice & 1.90 & 1.03 \\
            Full model & \textbf{1.39} & \textbf{0.58} \\
            \bottomrule
        \end{tabular*}
    \end{minipage}
\vspace{2mm}
    \caption{Ablations. (a) Geometry conditioning: Cross-attention shows the best performance among conditioning methods. (b) Tokenizer pathways: Both tokenizers contribute to GAIA's performance. Median relative $L^2$ error (\%).}
    \label{tab:ablations}
\end{table}
\vspace*{-6pt}
\section{Conclusion and Limitations}
\label{sec:limitations}
\vspace*{-6pt}
GAIA conditions integral kernels on tokenized representations of domain geometry to solve forward and  inverse problems on arbitrary domains in a single pass. It achieves state-of-the-art on all inverse and BVP benchmarks while maintaining stable accuracy across resolutions.
Three limitations bound the present work. First, GAIA is designed for time-independent operators; extending the framework to time-dependent problems requires additional architectural treatment for temporal evolution and is left to future work. Second, on forward problems where input and output share a fixed dense discretization, GAIA trades a small amount of in-distribution accuracy for substantially improved discretization invariance. Thirdly, the geometry-conditioning cross-attention scales as $\mathcal{O}(NK)$ in point-cloud size $N$ and token count $K$; for very large 3D point clouds ($N > 10^6$), latent compression or sparse attention \cite{katharopoulos2020transformers} may be needed to manage memory. 

\subsection*{Code and data availability.}
The code and newly introduced datasets are available at
\href{https://github.com/mkrishnan9/GAIA}{github.com/mkrishnan9/GAIA}
and \href{https://drive.google.com/drive/folders/1OSrAGvJh14M5APg-oRvTKb0MQYQgS6VR}{Google Drive},
respectively.

\section*{Funding}
This research used resources of the National Energy Research Scientific Computing Center (NERSC), a U.S. Department of Energy Office of Science User Facility, operated under Contract No. DE-AC02-05CH11231 using NERSC award ALCC-ERCAP0034775. MK and RD were partially supported by ONR Award N00014-23-1-2086. H.Y. was partially supported by the US National Science Foundation under awards IIS2520978, GEO/RISE-5239902, and the DARPA D24AP00325. Approved for public release; distribution is unlimited.




\bibliographystyle{unsrtnat}  
\bibliography{main}           

\appendix

\section{Datasets and Benchmarks} \label{app:datasets}

\begin{table}[hbt]
\centering
\small
\caption{Summary of benchmark problems. All datasets use median relative $L^2$ error (\%) as the evaluation metric.}
\label{tab:benchmarks}
\renewcommand{\arraystretch}{1.3}
\begin{tabularx}{\textwidth}{@{} p{0.2\textwidth} X p{0.33\textwidth} @{}}
\toprule
\textbf{Benchmark} & \textbf{Governing Equations \& Operator} & \textbf{Remarks} \\
\midrule

\textbf{Poisson-Gauss}~\citep{mousavi2025rigno}
&
$-\Delta u = f$, $x \in (0,1)^2$, $u|_{\partial\Omega}=0$.
Operator: $\mathcal{G}_{pg} : L^2(\Omega) \to H^1_0(\Omega) \cap H^2(\Omega)$, $f \mapsto u$.
Source $f$ is a superposition of $M$ random Gaussians.
&
\textbf{Forward.} Fixed unit-square domain. $128{\times}128$ uniform grid. 2048 train / 256 test samples.
\\

\textbf{Elasticity}~\citep{li2023geometry}
&
$\rho\,\partial_{tt} {u} - \nabla \cdot \boldsymbol{\sigma} = \mathbf{0}$.
Operator: $\mathcal{G}_{el} : C(\Omega) \ni d \mapsto \boldsymbol{\sigma} \in L^2(\Omega;\mathbb{R}^{2\times 2})$, where $d(x) = \mathrm{dist}(x,\Gamma_{\mathrm{inner}})$.
Incompressible Rivlin--Saunders constitutive model.
&
\textbf{Forward.} Unit square with randomly shaped interior hole ($r \in [0.2,\,0.4]$). 972 unstructured nodes. 1,024 train / 256 test samples.
\\

\textbf{3D Darcy Flow}
&
$-\nabla\cdot(\kappa( {x})\nabla u( {x})) = 1$, $u|_{\partial\Omega}=0$.
Operator: $\mathcal{G}_{df} : L^\infty(\Omega) \to H^1_0(\Omega)$, $\kappa \mapsto u$.
Log-normal permeability field; $\kappa = \exp(\gamma\,\mathcal{S}/\sigma_{\mathcal{S}})$.
&
\textbf{Forward.} Star-shaped 3D domains (spherical harmonic boundary, $l_{\max}=4$). Point cloud resolution 7000. 4096 train / 2048 test samples.
\\

\textbf{3D Poisson BVP}~\citep{yoo2025neural}
&
$-\Delta u = f_{\mathrm{analytical}}$, $u|_{\partial\Omega} = g(x,y,z)$.
Oscillatory analytical source; randomized polynomial Dirichlet BCs.
&
\textbf{BVP / Forward.} 3D mechanical parts (MCB): gears, nuts, fittings, screws \& bolts. Setting adapted from \cite{yoo2025neural}. 200 shapes/category train, 20 test with 50 boundary conditions per shape.
\\

\midrule

\textbf{EIT}~\citep{molinaro2023neural}
&
$\nabla\cdot(a( {x})\nabla u)=0$, mixed BCs.
Operator: $\Lambda_a : H^{1/2}(\partial\Omega) \to H^{-1/2}(\partial\Omega)$, $ u|_{\partial\Omega} \mapsto a\nabla u\cdot n$.
Inverse map: $\Lambda_a \mapsto a( {x})$.
&
\textbf{Inverse.} Extends~\citep{molinaro2023neural} to 2D star-shaped domains ($r(\theta)=r_0[1+c_1\cos 4\theta + c_2\cos 8\theta]$). $L=20$ Dirichlet patterns; $M=272$ boundary sensors. Log-trigonometric conductivity.
\\

\textbf{Optical Tomography}~\citep{fan2019solving}
&
$ {v}\cdot\nabla_x u + \sigma_t u = \sigma_s \int \Phi( {v}\cdot {v}')u\,d {v}'$.
Albedo operator: $\Lambda : L^1(\partial\Omega)\to L^1(\partial\Omega)$, $\varphi \mapsto J_+$.
Inverse map: Albedo difference data  $\Delta\Lambda \mapsto \delta\sigma_s( {x})$ scattering perturbation.
&
\textbf{Inverse.} Extends~\citep{fan2019solving} to convex 
pentagon domains. $S_{32}$ Discrete Ordinates solver; HG phase function ($g=0.9$). $32$ sources $\times$ $32$ receivers. 
\\
\textbf{Airfoil}~\citep{li2023fourier}
&
Compressible Euler equations around a 2D airfoil 
($M_\infty = 0.8$, $\text{AoA} = 0^\circ$).
Inverse map: sparse noisy Mach observations 
$\{\tilde{u}( {x}_j)\}_{j \in \mathcal{O}} \mapsto u( {x})$ 
over the full domain.
&
\textbf{Inverse.} Varying NACA airfoil geometries. 
$221 \times 51$ C-grid. $1\%$ relative Gaussian noise, $10\%$ sparse observations. 
1,000 train /  100 test.
\\

\bottomrule
\end{tabularx}
\end{table}

Below is a brief summary of the datasets that we use in the paper. 

\subsection{Inverse Problem: Electrical Impedance Tomography }
\label{ssec:eit}

Electrical Impedance Tomography (EIT) is an imaging technique that involves solving an inverse problem to reconstruct the internal material distribution of an object using boundary measurements. It is governed by the electrostatic equations with variable conductivity given by the Laplace equation with mixed Dirichlet and Neumann boundary conditions:
\begin{align}
    &\nabla \cdot (a(x) \nabla u(x)) = 0, \text{ for } x \in \Omega \subseteq \mathbb{R}^d,\ d= 2,3, \label{eqn:laplace}. 
\end{align}
Here, $a(x)$ is the conductivity distribution of the medium. The goal is to recover this distribution using only the boundary data on $\partial \Omega$ using the Dirichlet-to-Neumann map (DtN map, also known as the Calderon operator \cite{calderon2006inverse}) $\Lambda_a$ given by:
\[\Lambda_a:   H^{1/2}(\partial \Omega) \ni u_{|\partial \Omega}   \mapsto a \nabla u_{|\partial \Omega} \cdot n \in  H^{-1/2}(\partial \Omega).\]
where $H^{-1/2}(\partial \Omega)$ is the space of bounded linear functionals on the Sobolev space $H^{1/2}(\partial \Omega)$. In the case of full boundary measurement, it is a known result \citep{calderon2006inverse} that $a$ is uniquely determined under suitable assumptions. However, in practice, we only have a finite amount of noisy data, making the EIT problem highly ill-posed and sensitive to noise \citep{lionheart2004eit}.

\begin{figure}
    \centering
    \begin{subfigure}{0.58\linewidth}
        \centering
        \includegraphics[width=\linewidth, trim={4cm 0  4cm 0},clip]{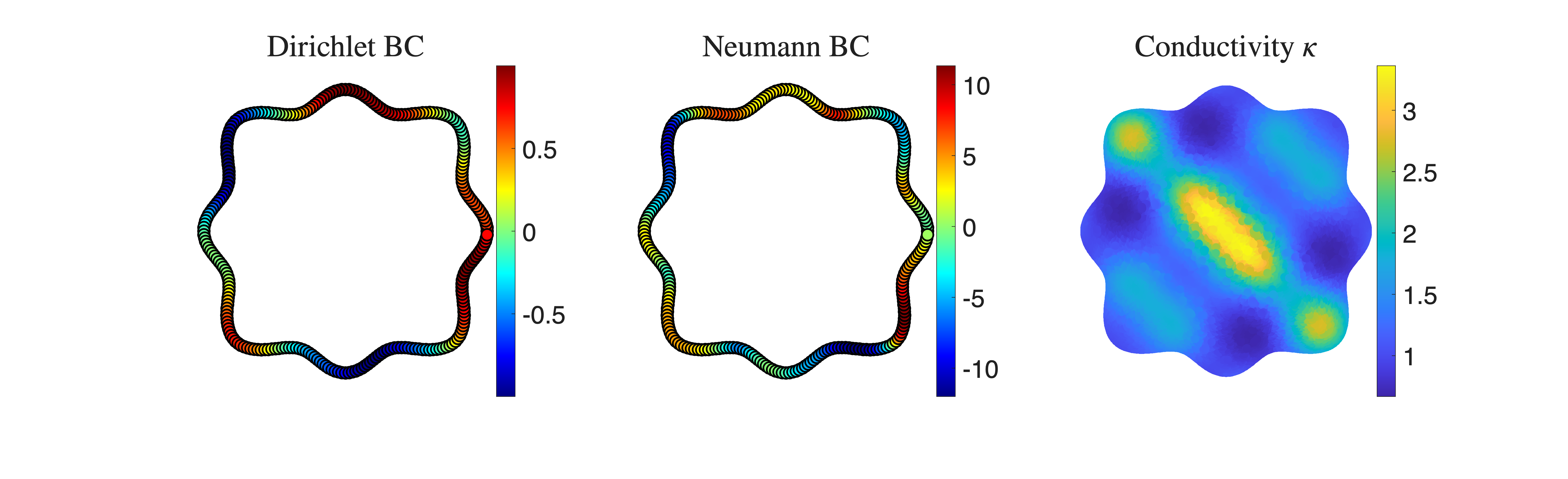}
        \caption{EIT: Dirichlet-Neumann boundary measurements.}
        \label{fig:calderon1}
    \end{subfigure}
    \hfill
    \begin{subfigure}{0.39\linewidth}
        \centering
        \includegraphics[width=\linewidth, trim={4cm 0  3cm 0},clip]{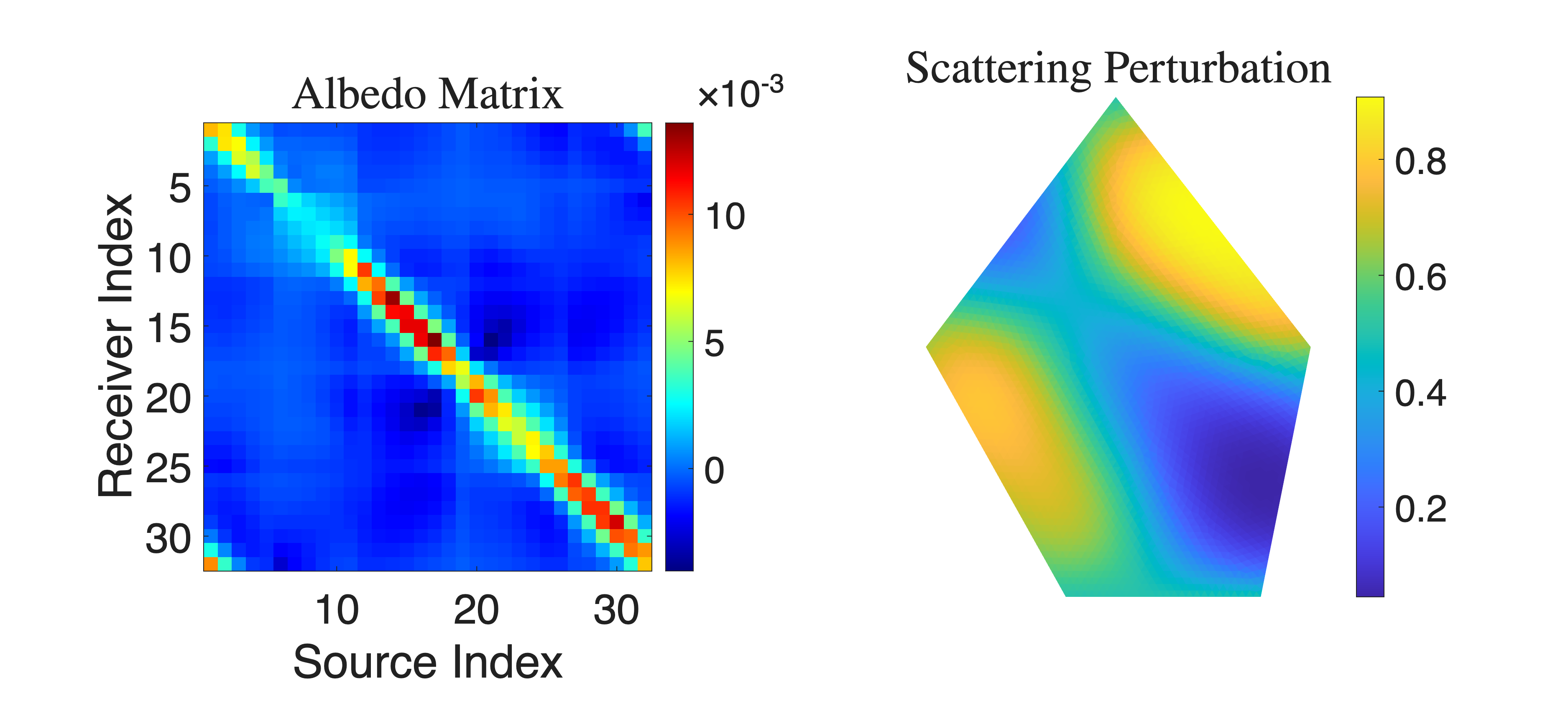}
        \caption{OT: Albedo matrix and scattering.}
        \label{fig:rte2}
    \end{subfigure}
    \caption{Inverse problems examples. (a) The EIT problem aims to reconstruct the internal conductivity distribution from boundary Dirichlet and Neumann values. (b) In OT, we reconstruct the scattering coefficient from the intensity at receivers given different source positions.} \vspace{-1mm}
    \label{fig:calderon}
\end{figure}

We aim to directly learn the inverse map using boundary measurements across different geometries. The training and test datasets were generated by solving \eqref{eqn:laplace} 
on strictly star-shaped domains $\Omega$ defined by the polar radius $r(\theta) = r_0[1 + c_1\cos(4\theta) + c_2\cos(8\theta)]$. Extending the data generation process in \cite{molinaro2023neural}, the isotropic conductivity coefficient $a(x,y)$ was drawn from a log-trigonometric probability distribution to ensure strict positivity and smoothness. Specifically, $a(x, y)$ is constructed as:
$$
a(x, y) = \exp\left(\sum_{k=1}^{m} c_k \sin(k\pi x) \sin(k\pi y)\right),
$$
where $m$ is drawn uniformly from $\{1, \dots, 5\}$, and the coefficients $\{c_k\}_{k=1}^m$ are sampled independently from a uniform distribution $\mathcal{U}([-1, 1])$. For each sampled conductivity, the governing equation was solved numerically using quadratic Finite Element Method (FEM) under $L=20$ distinct Dirichlet boundary conditions. The boundary data $\{g_\ell\}_{\ell=1}^{L}$ corresponds to plane waves incident from equispaced angles $\theta_\ell = \frac{2\pi (\ell-1)}{L}$, formulated as:
$$
u(x,y)\big|_{\partial \Omega} = g_\ell(x,y) = \cos\left(2\pi(x \cos \theta_\ell + y \sin \theta_\ell)\right).
$$
The corresponding Neumann measurement, $\Psi_\ell = a \frac{\partial u}{\partial \nu}$, representing the current flux, was evaluated at $M=272$ boundary sensors for each excitation $\ell$. Figure \ref{fig:calderon} shows the Dirichlet and Neumann measurements on the boundary for a given star-shaped domain and its corresponding recovered internal conductivity distribution.

\begin{figure}
    \centering
    \includegraphics[width=\textwidth]{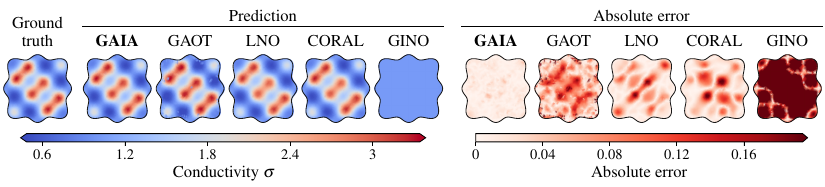}
    \caption{Model predictions on EIT conductivity reconstruction.}
    \label{fig:eit_predictions_app}
\end{figure}

\subsection{Inverse Problem: Optical Tomography}
\label{ssec:ot}

In optical tomography, the transport of photons through a scattering and absorbing medium occupying a bounded domain $\Omega \subset \mathbb{R}^d$ is modeled by the stationary radiative transport equation (RTE). Let $u(x, v)$ denote the particle density at spatial location $x \in \Omega$ and velocity direction $v \in S^{d-1}$. The medium is characterized by its scattering coefficient $\sigma_s(x) \ge 0$ and absorption coefficient $\sigma_a(x) \ge 0$. The RTE is given by:
\begin{align}
    v \cdot \nabla_x u(x, v) + \sigma_t(x)u(x, v) &= \sigma_s(x) \int_{S^{d-1}} \Phi(v \cdot v')u(x, v')dv', \quad x \in \Omega, \\
u(x, v) &= \phi(x, v), \quad (x, v) \in \Gamma_-.
\end{align}
Here, $\sigma_t(x) = \sigma_a(x) + \sigma_s(x)$ is the total attenuation coefficient, and $\Phi(v \cdot v')$ is the scattering phase function, normalized such that $\int_{S^{d-1}} \Phi(v \cdot v') dv' = 1$. The boundary phase space is partitioned into inflow ($\Gamma_-$) and outflow ($\Gamma_+$) boundaries, defined as 
\begin{equation}
    \Gamma_\pm = \{(x, v) \in \partial \Omega \times S^{d-1} : \pm n_x \cdot v > 0\},
\end{equation}
where $n_x$ is the unit outer normal vector at $x \in \partial \Omega$. In our experiment, following \cite{fan2019solving}, both the boundary illumination and the recorded data are assumed to be independent of the velocity variable. Specifically, the prescribed incoming particle density is modeled as an isotropic source depending only on the spatial variable, $\phi(x) \in L^1(\partial \Omega)$. Similarly, the measurements on the outflow boundary record the outgoing angular flux (or current), integrating the particle density over all outward-pointing velocities: $J_+(x) = \int_{v \cdot n_x > 0} (v \cdot n_x) u(x, v) dv$. The associated boundary observation operator, known as the Albedo operator $\Lambda$, thus maps the spatial input on the boundary to the spatial output:
\begin{equation}
    \Lambda : L^1(\partial \Omega) \to L^1(\partial \Omega), \quad \Lambda : \phi(x) \mapsto J_+(x).
\end{equation}
The continuous inverse problem seeks to recover the unknown medium properties, characterized by the scattering and absorption coefficients, from the knowledge of $\Lambda$. Formally, this defines the continuous inverse map $\mathcal{F}^{-1}$ from the space of bounded linear operators to the space of continuous medium parameters:$$\mathcal{F}^{-1} : \mathcal{L}\left(L^1(\partial \Omega), L^1(\partial \Omega)\right) \to C(\Omega) \times C(\Omega), \quad \mathcal{F}^{-1}(\Lambda) = (\sigma_s, \sigma_a).$$The well-posedness and Lipschitz stability of this inverse map have been established under suitable conditions \citep{bal2010inverse}. 

For the numerical experiments and data generation in this work, we restrict our domain $\Omega \subset \mathbb{R}^2$ to randomly scaled convex pentagons. We assume the absorption coefficient is a known, spatially invariant constant, $\sigma_a(x) \equiv 0.01$, reducing the inverse problem to the recovery of the spatially varying scattering coefficient $\sigma_s(x)$. The highly forward-peaked anisotropic scattering in biological tissues is modeled using the 2D Henyey-Greenstein (HG) phase function,$$\Phi(v \cdot v') = \frac{1 - g^2}{2\pi(1 + g^2 - 2g(v \cdot v'))},$$  
where the anisotropy factor is set to $g = 0.9$. The spatial distribution of the scattering coefficient is constructed using a randomized superposition of low-frequency trigonometric modes passed through a scaled logistic sigmoid function. Let $x' = (x'_1, x'_2)$ denote the spatial coordinates normalized by the domain scale factor. The scattering field is defined as:
$$\sigma_s(x) = 1.0 + \left({1 + \exp\left(-3 \sum_{k=1}^m c_k \sin(\omega_{x,k} \pi x'_1) \sin(\omega_{y,k} \pi x'_2)\right)}\right)^{-1},$$
where the number of modes $m$ is drawn uniformly from $\{2, \dots, 5\}$, the expansion coefficients $c_k$ are sampled independently from a uniform distribution $\mathcal{U}([-1, 1])$, and the spatial frequencies $\omega_{x,k}$ and $\omega_{y,k}$ are drawn uniformly from the integer set $\{1, 2, 3\}$. This construction ensures that the scattering coefficients are smooth and strictly bounded within the physical range $\sigma_s(x) \in (1.0, 2.0)$.

The continuous boundary illumination is discretized using $N_{src} = 32$ localized isotropic point sources distributed uniformly along $\partial \Omega$. The forward transport problem is solved using the Discrete Ordinates ($S_{32}$) method with angular discretization of 32 angles, and a Finite Element spatial discretization, using GMRES for the source iteration. The corresponding outgoing current is measured at $N_{rec} = 32$ interleaved receiver locations. This discrete setup approximates the Albedo operator as a dense matrix $\Lambda \in \mathbb{R}^{32 \times 32}$. To isolate the scattering anomalies and suppress structural geometric artifacts, we employ a difference imaging approach. A baseline Albedo matrix $\Lambda_0$ is simulated using a known, constant background scattering coefficient $\sigma_{s,0} = 1.0$. The objective of the discrete inverse problem is then to infer the scattering perturbation $\delta \sigma_s(x) = \sigma_s(x) - \sigma_{s,0}$ strictly from the measured difference data $\Delta \Lambda = \Lambda - \Lambda_0$.

\begin{figure}[!htb]
    \centering
    \includegraphics[width=\textwidth]{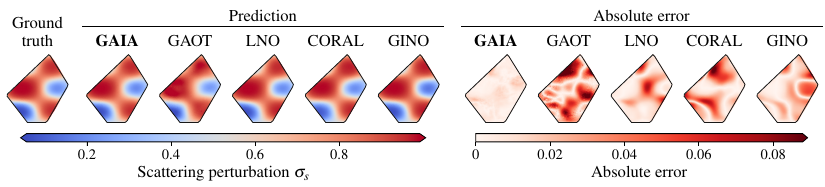}
    \caption{Model predictions on RTE. Comparison of ground truth solution against predictions from GAIA, GAOT, LNO, CORAL, and GINO.}
    \label{fig:rte_predictions}
\end{figure}

\subsection{Inverse Problem: Airfoil Flow Reconstruction}
\label{ssec:airfoil}

We consider the problem of reconstructing a full aerodynamic flow field from sparse, noisy 
measurements over a family of transonic solutions over airfoil geometries. The underlying physical system 
is governed by the compressible Euler equations,
\begin{equation}
    \frac{\partial \rho_f}{\partial t} + \nabla \cdot (\rho_f  {v}) = 0, \qquad
    \frac{\partial \rho_f  {v}}{\partial t} + \nabla \cdot (\rho_f  {v} \otimes  {v} + p {I}) = 0, \qquad
    \frac{\partial E}{\partial t} + \nabla \cdot \bigl((E + p) {v}\bigr) = 0,
\end{equation}
where $\rho_f$ is the fluid density, $ {v}$ is the velocity vector, $p$ is the pressure, 
and $E$ is the total energy. Viscous effects are neglected. Far-field conditions are prescribed 
as $\rho_\infty = 1$, $p_\infty = 1.0$, $M_\infty = 0.8$, ${AoA} = 0^\circ$, with a 
no-penetration condition imposed at the airfoil surface. Here $M_\infty$ is the Mach number and ${AoA} $ is the angle of attack.

The dataset is drawn from the NACA airfoil corpus of~\cite{li2023fourier}, comprising 
$N = 1{,}200$ flow solutions, one per airfoil shape, generated with a second-order implicit 
finite volume solver on a body-fitted C-grid of $N_x \times N_y = 221 \times 51 = 11{,}271$ 
quadrilateral elements, refined near the airfoil surface. Airfoil geometries are parameterized 
via a design element approach~\cite{farin2014curves} in which the baseline NACA-0012 profile 
is deformed by displacing the control nodes of an enclosing cubic design element in the 
vertical direction, with displacements drawn from $d \sim \mathcal{U}[-0.05, 0.05]$. The 
dataset is partitioned into $N_{\mathrm{tr}} = 1{,}000$ training, $N_{\mathrm{val}} = 100$ 
validation, and $N_{\mathrm{te}} = 100$ test configurations.

We formulate it as a 
sparse-to-full field reconstruction problem. The quantity of interest is the steady-state 
Mach number field $u : \Omega \to \mathbb{R}$, discretized over all $11{,}271$ mesh nodes. 
At inference time, only a sparse subset of observations 
$\{u( {x}_j)\}_{j \in \mathcal{O}}$, with 
$\mathcal{O} \subset \{1, \ldots, N_x N_y\}$, is available as input, and these observations 
are further corrupted by zero-mean Gaussian noise scaled to the per-sample peak Mach number:
\begin{equation}
    \tilde{u}_i( {x}_j) \;=\; u_i( {x}_j) \;+\; \eta_{i,j}, \qquad
    \eta_{i,j} \sim \mathcal{N}\!\left(0,\, \sigma^2 \|u_i\|_\infty^2\right), \quad j \in \mathcal{O},
    \label{eq:obs-noise}
\end{equation}
with $\sigma = 0.01$, corresponding to $1\%$ of the per-sample $L^\infty$ norm. The task 
is to recover the complete Mach field $u$ at all grid nodes, including those with no sensor 
coverage. The 
model must simultaneously denoise, interpolate, and extrapolate from incomplete observations 
over geometrically varying domains.

\begin{figure}
    \centering
    \includegraphics[width=\textwidth]{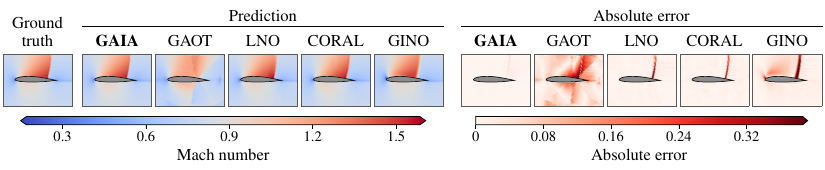}
    \caption{Model predictions on Airfoil Mach field reconstruction from sparse noisy observations.}
    \label{fig:inverse_predictions_app}
\end{figure}

\subsection{3D Boundary Value Problem: Poisson on MCB}
\label{ssec:mcb}

To test the generalization of our framework to unseen problem domains, we construct a PDE dataset using the Mechanical Components Benchmark (MCB) dataset \citep{kim2020large}, which contains a variety of 3D mechanical part shapes. We adapt the tetrahedral meshes for shapes in four categories (SCREWS \& BOLTS, NUTS, FITTINGS, and GEARS) in \cite{yoo2025neural} built by meshing the interiors of unit-cube-normalized shapes using fTetWild \citep{hu2020fast}.  We used a similar setting as in \cite{yoo2025neural} but formulated and generated data for a boundary value problem. The shape collection is divided into 200 shapes for training and 20 shapes for testing per category. Shapes within the same category exhibit diverse geometries, presenting a standard challenge for learned solution operators to generalize effectively.

We define a highly non-linear, oscillatory analytical source term $f_{\text{analytical}}(x,y,z)$. The base function is constructed using trigonometric variations along all three spatial axes:$$f_{\text{analytical}} = T_1(x,y) + T_2(x,y) + T_3(z)$$where the individual components are defined by the constants $A=1.25$, $B=1.5$, $C=1.5$, and $D=1.5$, such that:$$T_1(x,y) = -\left((A\pi)^2 + (AC\pi)^2\right) \sin(A\pi x) \cos(AC\pi y)$$$$T_2(x,y) = (A\pi)^2 \cos(A\pi x)\left(1 - \sin(AB\pi y)\right) + (AB\pi)^2 \sin(AB\pi y)\left(1 - \cos(A\pi x)\right)$$$$T_3(z) = 2(AD\pi)^2 \cos(2AD\pi z).$$

To match the magnitude of the source fields released by \cite{yoo2025neural}, we use the affinely
rescaled source $f = \alpha_c f_{\text{analytical}} + \beta_c$, where the constants
$(\alpha_c, \beta_c)$ for each category $c$ are obtained by a least-squares fit of
$f_{\text{analytical}}$ to the released source field on a reference shape of that category. These constants are held fixed for all shapes and samples
within a category (\textsc{Fittings}: $(-0.8641, 15.8068)$; \textsc{Gears}: $(-0.8463, 12.2468)$;
\textsc{Nuts}: $(-0.7511, -1.0794)$; \textsc{Screws \& Bolts}: $(-1.7921, 121.6870)$).

To generate multiple unique samples per geometry, we randomize the Dirichlet boundary conditions $g(x,y,z)$ at the mesh surface. The boundary values are prescribed using a parameterized polynomial function:$$g(x,y,z) = E(x^3 - 3xy^2) + F(y^3 - 3x^2y) + (x^2 - z^2)$$For each generated sample, the coefficients $E$ and $F$ are sampled from uniform distributions such that $E \sim \mathcal{U}(-1.0, 1.0)$ and $F \sim \mathcal{U}(0.0, 1.0)$.

To generate the dataset, we employ the FEniCS finite element framework \citep{logg2012automated} to solve Poisson’s equation on the generated meshes.

\begin{figure}
    \centering
    \includegraphics[width=\linewidth, trim={0 3cm  0 0},clip]{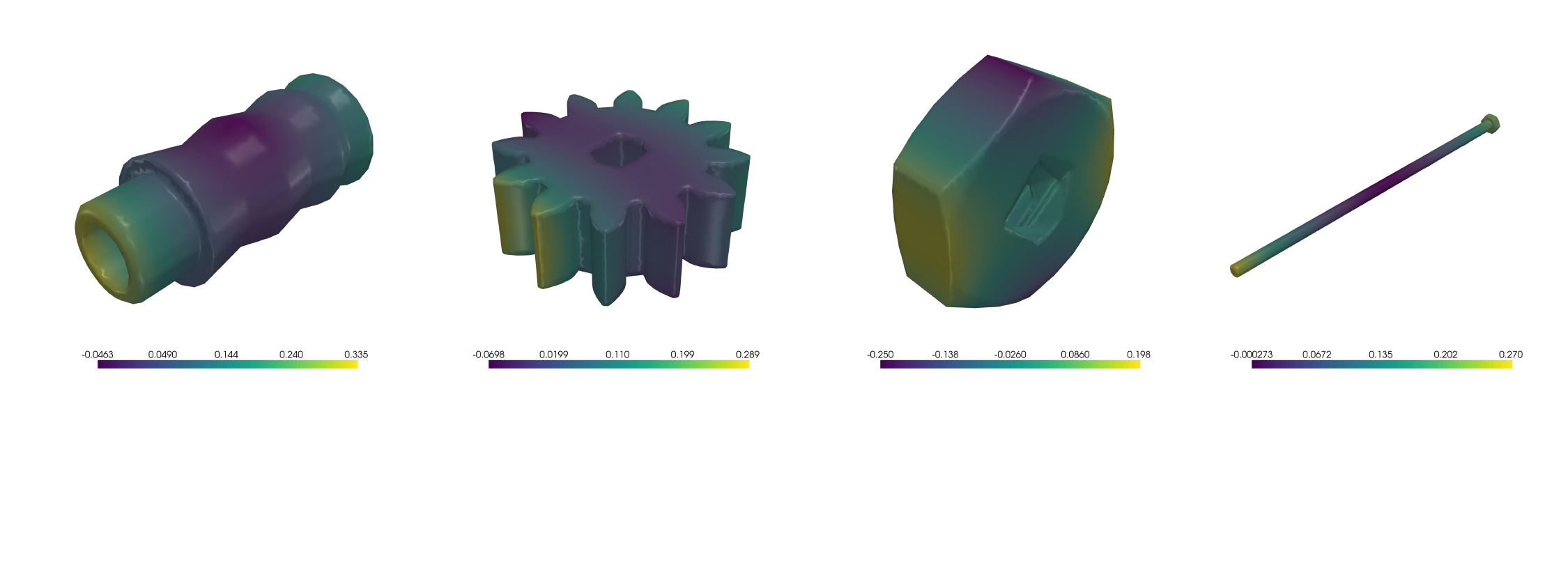}
    \caption{Boundary Values on MCB shapes}
    \label{fig:mcb_bdry}
\end{figure}

\begin{figure}
    \centering
    \includegraphics[width=\textwidth, trim=0 3mm 0 0cm, clip]{ 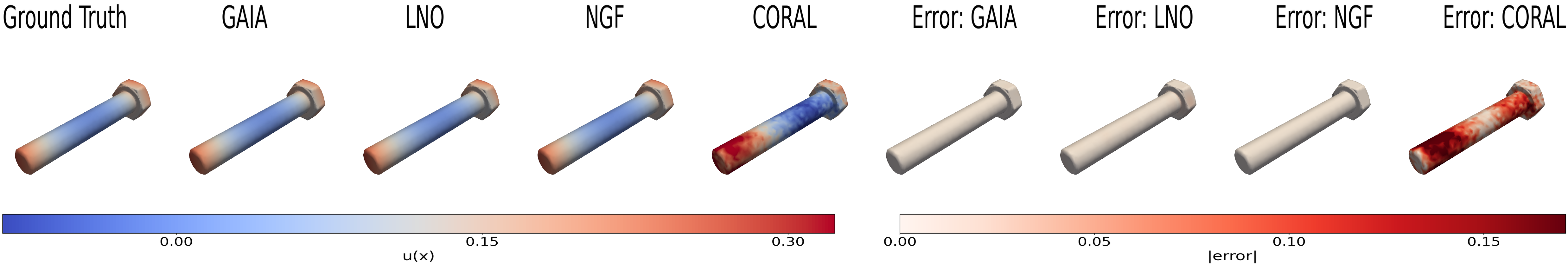}
    \caption{Results on the screws\&bolts category from the Mechanical Components Benchmark.}
    \label{fig:screws}
\end{figure}

\subsection{Forward Problem: 3D Darcy Flow}
\label{ssec:darcy}

We consider the stationary Darcy flow through a heterogeneous porous medium. Let the spatial domain $\Omega \subset \mathbb{R}^3$ represent  three-dimensional domain. The fluid pressure $u( {x})$ is modeled by the elliptic partial differential equation:
$$-\nabla \cdot (\kappa( {x}) \nabla u( {x})) = 1, \quad  {x} \in \Omega,$$
subject to homogeneous Dirichlet boundary conditions, $u|_{\partial \Omega} = 0$. Here, $\kappa( {x})$ represents the strictly positive permeability field of the medium. The operator learning objective is to approximate the continuous solution operator $\mathcal{G}_{d}$ that maps the permeability field, defined on a variable geometries, to the resulting scalar pressure field. Formally, for a given domain $\Omega$, the mapping is defined as:
$$\mathcal{G}_{df} : L^\infty(\Omega) \to H_0^1(\Omega), \quad \mathcal{G}_{df}(\kappa) = u.$$

To learn this operator, we construct a dataset by solving the Darcy flow equation on $\Omega \subset \mathbb{R}^3$. The boundary $\partial \Omega$ of each sample is generated as a "star shape" defined by a random spherical harmonic expansion up to degree $l_{max}=4$:
\begin{equation}
    r(\theta, \phi) = \sum_{l=0}^{4} \sum_{m=-l}^{l} c_{lm} Y_l^m(\theta, \phi), \quad c_{lm} \sim \mathcal{N}(0, \sigma_l).
\end{equation}
The resulting geometries are discretized into watertight unstructured tetrahedral meshes. The permeability field $\kappa( {x})$ is computed as a continuous log-normal random field, $\kappa( {x}) = \exp(\gamma \cdot \mathcal{S}( {x}) / \sigma_{\mathcal{S}})$, where $\mathcal{S}( {x})$ is a superposition of $M=30$ random Fourier modes:
\begin{equation}
    \mathcal{S}( {x}) = \sum_{j=1}^{M} \alpha_j \cos( {k}_j \cdot  {x} + \phi_j).
\end{equation}
Here, the wave vectors $ {k}_j \sim \mathcal{N}( {0}, \sigma_k^2  {I})$ control the spatial correlation length, phases $\phi_j \sim \mathcal{U}[0, 2\pi]$ ensure translation invariance, and amplitudes $\alpha_j \sim \mathcal{N}(0, 1)$ provide variance. The field is normalized by its sample standard deviation $\sigma_{\mathcal{S}}$ and scaled by $\gamma=0.5$ to ensure a smooth, strictly positive, and spatially heterogeneous medium.

The forward problem is solved using the standard Galerkin Finite Element Method (FEM) implemented in FEniCS. We utilize linear Lagrange ($P_1$) basis functions for both the solution space and the coefficient field. Following the FEM solution, we extract the training data by randomly sampling $N_p=7000$ points within the domain volume for each simulation.


\begin{figure}[!htb]
    \centering
    \begin{subfigure}{0.49\linewidth}
        \centering
        \includegraphics[width=\linewidth, trim={2cm 0  2cm 0},clip]{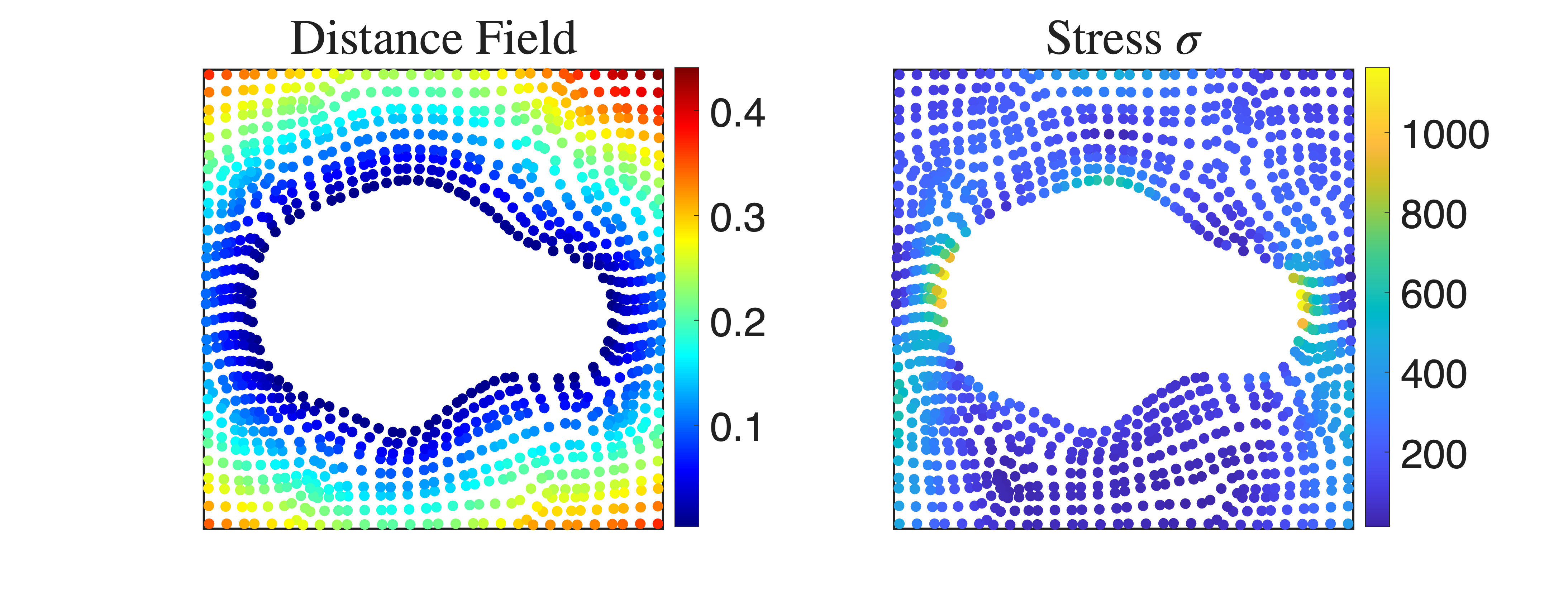}
        \caption{Elasticity}
        \label{fig:elastic}
    \end{subfigure}
    \hfill
    \begin{subfigure}{0.49\linewidth}
        \centering
        \includegraphics[width=\linewidth,trim={2cm 0  2cm 0},clip]{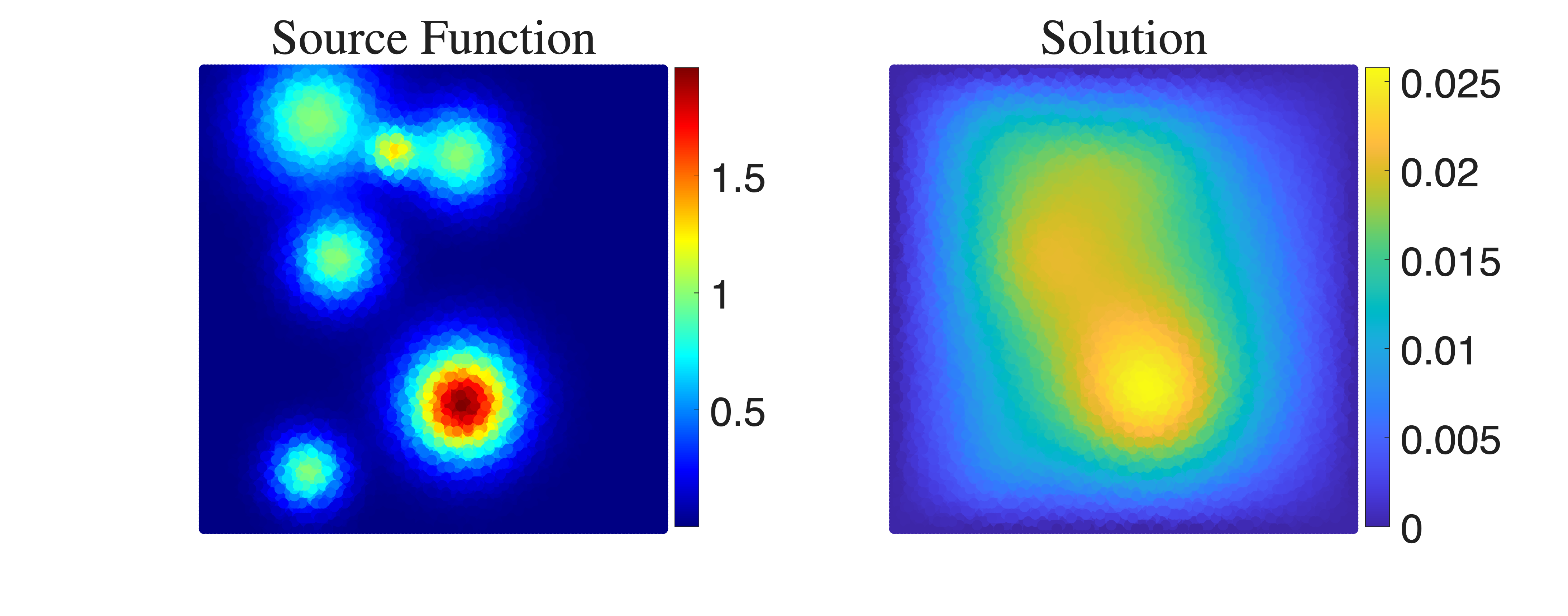}
        \caption{Poisson-Gauss}
        \label{fig:pg}
    \end{subfigure}
    \caption{Input output fields}
    \label{fig:elas_pg}
\end{figure}

\subsection{Forward Problem: Elasticity Problem}
\label{ssec:elas}

We evaluate the model on a standard Elasticity benchmark, where the objective is to learn the mapping from a domain's geometry to its resulting internal stress field under external loading. The general force balance of a solid body is governed by the hyperelastic equation:
$$\rho \partial_{tt}  {u} - \nabla \cdot \boldsymbol{\sigma} =  {0},$$
where $\rho$ represents the mass density, $ {u}$ the displacement field, and ${\sigma}$ the Cauchy stress tensor. The strain field is related to the displacement field via standard kinematic relations, and the system is closed by a hyperelastic constitutive model describing the non-linear stress-strain relationship. For this benchmark from \cite{li2023fourier}, the stationary solution is considered for a unit square hyper-elastic incompressible Rivlin-Saunders specimen with a hole at its center. The geometry of the hole is randomly sampled such that the radius always takes a value between $0.2$ and $0.4$ (see \cite{li2023fourier} for details). The specimen is subjected to mixed boundary conditions: it is clamped at the bottom boundary ($ {u} = 0$) and is under a constant vertical tension traction on its top boundary. The stress field is computed for different geometries with a finite elements solver with about $100$ quadratic quadrilateral elements. The target function in this dataset is the stress field $\sigma$, which is available at $972$ unstructured coordinates. The operator learning task thus seeks to approximate the continuous solution operator $\mathcal{G}_{el}$ that maps the geometry of the domain directly to the stress field. The geometry is encoded as the shortest-path distance function from the inner boundary of the hole, $d(x) = \text{dist}(x, \Gamma_{\text{inner}})$. The formal operator mapping is thus defined as:$$\mathcal{G}_{el} : C(\Omega) \ni d \to \boldsymbol{\sigma} \in L^2(\Omega; \mathbb{R}^{2 \times 2}).$$

Figure \ref{fig:elastic} demonstrates an example of a sample with a deformation in the center. The input field is distance of the unordered point cloud from the inner boundary.

\begin{figure}[!htb]
    \centering
    \includegraphics[width=\textwidth]{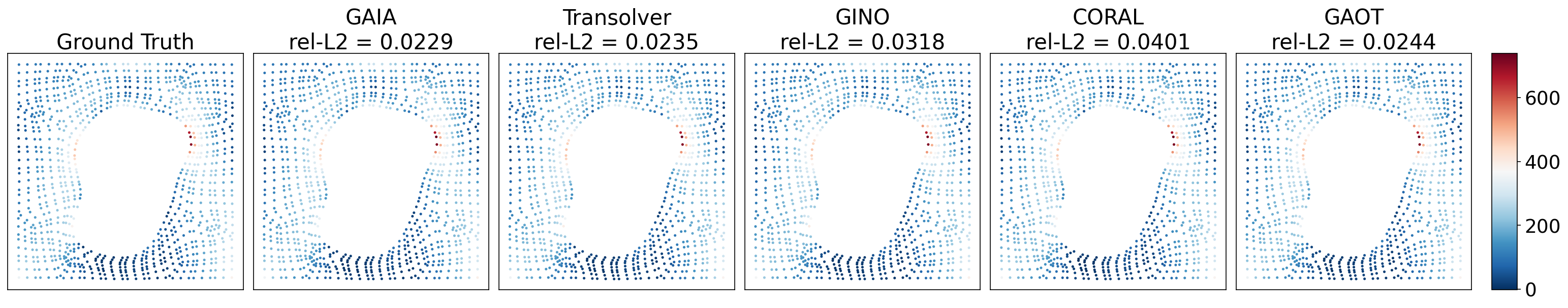}
    \caption{Model predictions on Elasticity. Comparison of ground truth solution against predictions from GAIA, Transolver, CORAL, GINO and GAOT.}
    \label{fig:elas_predictions}
\end{figure}

\subsection{Forward Problem: Poisson-Gauss}
\label{ssec:pg}

To demonstrate GAIA's performance on a standard fixed-grid benchmark, we consider the classical Poisson's equation \cite{mousavi2025rigno}. Let the domain be the unit square $\Omega = (0,1)^2$. The governing PDE is given by:
\begin{equation}
    -\Delta u(x) = f(x),\ x \in (0,1)^2 ,
\end{equation}
subject to homogeneous Dirichlet boundary conditions, $u|_{\partial \Omega} = 0$. The objective of the neural operator is to learn the inverse Laplacian operator $\mathcal{G}_{pg} = -\Delta^{-1}$, which maps the right-hand side source function to the scalar solution field. Formally, this defines the continuous solution operator:
\begin{equation}
    \mathcal{G}_{pg} : L^2(\Omega) \to H_0^1(\Omega) \cap H^2(\Omega), \quad \mathcal{G}_{pg}(f) = u.
\end{equation}
The source term $f$ is procedurally generated as a superposition of $M$ random Gaussian pulses:
\begin{equation}
f(x, y) = \sum_{i=1}^{M} \exp\left( -\frac{(x - \mu_{x,i})^2 + (y - \mu_{y,i})^2}{2\sigma_i^2} \right), 
\end{equation}
where $M$ is an integer drawn from a geometric distribution. The mean coordinates are sampled uniformly such that $\mu_{x,i}, \mu_{y,i} \sim \mathcal{U}(0,1)$, and the standard deviations are sampled as $\sigma_i \sim \mathcal{U}(0.025, 0.1)$. The ground truth data pairs $(f, u)$ are evaluated on a high-resolution $128 \times 128$ uniform grid. We refer to \cite{mousavi2025rigno} for further dataset details. Figure \ref{fig:pg} visualizes a representative input source function and its corresponding computed solution.

\begin{figure}
    \centering
    \includegraphics[width=\textwidth]{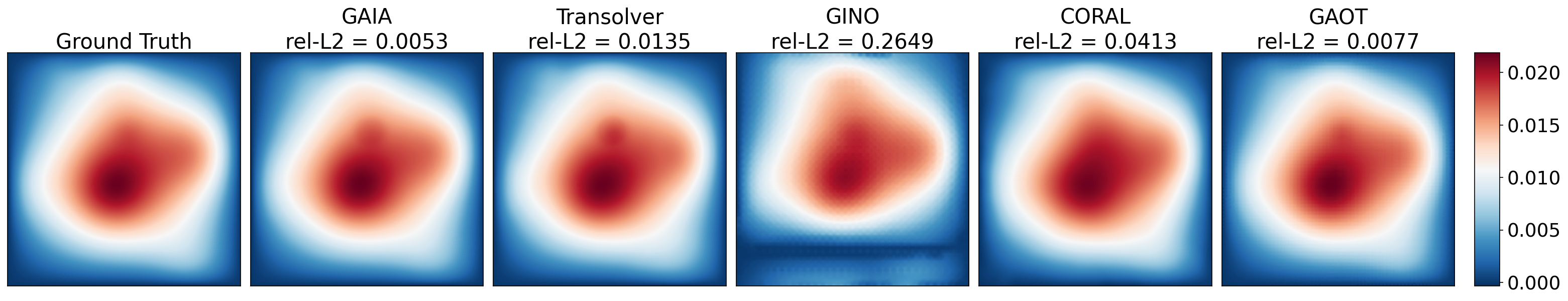}
    \caption{Model predictions on Poisson-Gauss. Comparison of ground truth solution against predictions from GAIA, Transolver, CORAL, GINO and GAOT.} \vspace*{-5pt}
    \label{fig:pg_predictions}
\end{figure}

\section{GAIA Implementation Details} \label{app:implementation_details}

All models were trained on a single NVIDIA A6000 GPU using the Adam optimizer~\cite{kingma2014adam} with cosine decay learning rate scheduling and 5\% warmup starting at 10\% of the peak learning rate. We use 8 attention heads, Fourier positional encodings with 10 frequency bands and base frequency $\omega_0 = 30$. All hyperparameters for experiments are given in Table \ref{tab:hyperparams}.

\begin{table}[h!]
\centering
\caption{Training and architectural hyperparameters across benchmarks. BS refers to batch size, ep.: epochs, $N_{\text{in}}$/$N_{\text{out}}$: number of input/output points per sample, GHD: Hidden dimension, GAD: Token dimension x Number of tokens,  AD: attention dimension, NGT: Number of geometric tokens.}
\label{tab:hyperparams}
\small
\setlength{\tabcolsep}{3pt}
\begin{tabular}{l|ccccccc|ccc|cccc}
\toprule
& \multicolumn{7}{c|}{Training} & \multicolumn{3}{c|}{Architecture} & \multicolumn{4}{c}{Geometry Conditioning} \\
\textbf{Dataset} & $N_{\text{train}}$ & $N_{\text{test}}$ & $N_{\text{in}}$ & $N_{\text{out}}$ & BS & LR & Ep. & Modes & Width & Blocks & GHD & GAD & AD & NGT. \\
\midrule
OT          & 8000 & 2000 & $32\times32$ & $\sim$5000    & 4 & 2e-4 & \multirow{6}{*}{500} & 96 & \multirow{6}{*}{128} & 4 & \multirow{5}{*}{128} & 64 & \multirow{3}{*}{128} & 8 \\
EIT         & 4096 & 2048 & 272          & 2000          & 4 & 1e-4 &                      & 32 &                      & 5 &                      & 32 &                      & 8 \\
Airfoil     & 1000 & 100  & 1127         & 11271         & \multirow{3}{*}{8} & \multirow{3}{*}{5e-4} &                      & 96 &                      & 4 &                      & 256 &                     & 4 \\
Elasticity  & 1024 & 256  & 972          & 972           &  &  &                      & \multirow{3}{*}{128} &                     & \multirow{3}{*}{5} &                      & \multirow{2}{*}{128} & \multirow{2}{*}{64}                  & 16 \\
P-Gauss     & 2048 & 256  & $128^2$      & $128^2$       &  &  &                   &  & 64                  &  &                   &  &                  & 8 \\
3D Darcy    & 4096 & 2048 & 7000         & 7000          &4  & 2e-4 &                      &  & 128                     &  & 256                  & 256 & 128                 & 8 \\
\midrule
MCB: Gear    & \multirow{4}{*}{10k} & \multirow{4}{*}{1k} & \multirow{4}{*}{$\sim$6700} & \multirow{4}{*}{$\sim$15800} & \multirow{4}{*}{1} & 2e-4 & \multirow{4}{*}{100} & \multirow{3}{*}{128} & \multirow{4}{*}{128} & \multirow{4}{*}{4} & \multirow{3}{*}{128} & \multirow{3}{*}{128} & \multirow{4}{*}{256} & \multirow{3}{*}{4} \\
MCB: Screw   & & & & & & 1e-4 & & & & & & & & \\
MCB: Fitting & & & & & & 1e-4 & & & & & & & & \\
MCB: Nut     & & & & & & 2e-4 & & 96 & & & 256 & 256 & & 8 \\
\bottomrule
\end{tabular}
\end{table}

\section{Baselines} \label{app:baselines}

Unless specifically noted, all baseline implementations follow their default hyperparameters in their published repositories. While CORAL and LNO can technically handle varying input/output spatial domains, the codebase had to be modified slightly to accomplish this.

\subsection{CORAL}
CORAL~\citep{serrano2023operator} casts operator learning as a regression in
the latent spaces of two implicit neural representations (INRs). Each input
sample $a_i$ and output sample $u_i$ is encoded by fitting a shift-modulated
SIREN via auto-decoding: a
shared base network is trained jointly with per-sample latent codes
$z_{a_i}, z_{u_i}$, where the codes are passed through small hypernetworks
that produce additive shifts on the SIREN activations. The codes are obtained
through a short inner loop of gradient descent on the reconstruction loss. Once the two INRs are trained, a separate network learns
the mapping $z_{a_i} \mapsto z_{u_i}$ in latent space. Because the INRs are
queried point-wise, CORAL natively supports varying input and output meshes
across samples.

We use the official implementation\footnote{\url{https://github.com/LouisSerrano/coral}}.
Both INRs are 4-layer SIRENs of hidden width 256 with shift-only modulation
generated by a single-layer hypernetwork of width 128. The latent regression
network $z_{a} \mapsto z_{u}$ is a 3-block residual MLP with Swish
activations, trained with Adam (no weight decay) and a step decay of
$\gamma=0.9$. The two-stage protocol trains the INRs first, then freezes them
and fits the regression network. We meta-learn the inner learning rate at a
meta-rate of $10^{-4}$. As the official codebase assumes a shared mesh
between input and output, we modified the data pipeline to query the two
INRs on distinct point clouds for boundary-value problems and inverse problems where sensor and reconstruction grids differ.

\subsection{GINO}
GINO~\citep{li2023geometry} is a hybrid neural operator that combines a graph
neural operator (GNO)~\citep{li2020neural} with a Fourier neural operator
(FNO) to handle inputs and outputs on irregular point
clouds. An input GNO encoder integrates the input function over local
neighborhoods of each latent grid point, lifting the irregular sample onto a
regular Cartesian latent grid. An FNO processor operates on this latent grid,
exploiting the FFT for global integration. An output GNO decoder then
projects the processed latent field back onto an arbitrary query point cloud.
The encoder and decoder rely on radius graphs: each output node attends to all
input nodes within a fixed radius $r$.

We use the official implementation\footnote{\url{https://github.com/neuraloperator/neuraloperator}}.
The FNO processor uses 6 layers with hidden width 32, group normalization,
and a channel MLP expansion of 0.5; AdaIN conditioning is enabled with 8
feature channels. Both GNO encoder and decoder use linear kernel transforms
with a half-cosine kernel weighting. The input GNO MLP has
hidden layers $[128, 256, 128]$ and the output GNO MLP has hidden layers
$[256, 512, 256]$.  The GNO radius $r$ is tuned to the characteristic length scale of each domain and is shared between the input and output GNO; values are listed in Table~\ref{tab:gino_hparams}.

\begin{table}[h]
\centering
\caption{GINO GNO radius $r$ across benchmarks. The same radius is used for
the input and output GNO.}
\label{tab:gino_hparams}
\small
\begin{tabular}{lcccccc}
\textbf{Dataset} & EIT & Airfoil & Poisson-Gauss & Elasticity & RTE & 3D Darcy \\
\midrule
$r$ & 0.25 & 0.125 & 0.125 & 0.125 & 5.0 & 0.25 \\
\end{tabular}
\end{table}

\subsection{Transolver}
Transolver~\citep{wu2024transolver} is a transformer-based neural operator
that adapts standard self-attention to PDE solving on irregular meshes.
Rather than computing attention directly between mesh nodes -- which scales
quadratically with the point-cloud size -- it introduces a Physics-Attention
mechanism that (softly) assigns each mesh point to a small set of learnable
physical states (or \emph{slices}), runs self-attention over the slice
representations, and then projects back to the original mesh. This reduces
attention cost from $\mathcal{O}(N^2)$ to $\mathcal{O}(N K)$ for $K$ slices
and lets the model scale to large unstructured meshes. Because both the
slice assignment and the back-projection share the same point set,
Transolver requires the input and output to live on a common discretization,
which is why we omit it from inverse and BVP comparisons.

We use the official implementation\footnote{\url{https://github.com/thuml/Transolver}}.
The model uses 8 Transolver blocks of hidden width 256 with 8 attention
heads, an MLP expansion ratio of 2, and 32 slice tokens per block; the slice
projection uses 8 reference points. Training uses AdamW with a learning rate of
$10^{-3}$, weight decay $10^{-5}$, and batch size 20.

\subsection{Geometry Aware Operator Transformer (GAOT)}
GAOT~\citep{wen2025geometry} is a graph-based transformer architecture for
operator learning on unstructured grids. It uses a multiscale graph encoder
(MAGNO) to aggregate point-cloud features onto a fixed Cartesian latent
token grid, applies transformer attention over the latent tokens, and
decodes back to the mesh through a corresponding graph decoder. 


For the forward problems (Elasticity, Poisson-Gauss and 3D Darcy), we adopt the default hyperparameters provided in the GAOT repository without modification \footnote{\url{https://github.com/camlab-ethz/GAOT}}. For 2D problems, the latent token grid is 64×64; for the 3D Poisson problem, we use a 64×32×32 latent grid as specified in the original paper \cite{wen2025geometry}. Training uses the default AdamW optimizer with a mixed learning rate schedule (linear warmup followed by cosine decay).

For the three inverse problems (EIT, airfoil flow and OT), we use a modified version of the GAOT codebase. Since inverse problems involve distinct input and output domains (e.g., boundary measurements to interior fields), we modify the MAGNO graph construction to build separate encoder and decoder neighborhood graphs, rather than a single shared graph as in the original implementation. We additionally lower the initial learning rate to $1e^{-4}$, as the default $8e^{-4}$ did not yield good performance on this problem. All other architectural hyperparameters remain at their defaults. 

\subsection{Latent Neural Operator (LNO)}
LNO~\citep{wang2024latent} learns operator mappings via attention through a low-dimensional latent space. The encoder applies cross-attention between a fixed set of learnable latent queries and the input point cloud, producing a compact latent representation. A stack of self-attention blocks then processes these latent tokens. Finally, the decoder applies cross-attention from the output query points to the processed latent tokens to produce the
predicted field. 

We apply LNO to the four forward Poisson problems on MCB dataset, using the Plasticity config from the original repository\footnote{\url{https://github.com/L-I-M-I-T/LatentNeuralOperator}} as the reference. The architectural hyperparameters are kept identical. Batch size is reduced to 1 to accommodate the MCB meshes in memory, and the model is trained for 100 epochs like our models. For the inverse problems, we use the LNO propagator config from the original repository as the reference, with batch size set to 64 for speed.

\subsection{Neural Greens Function (NGF)}
NGF~\citep{yoo2025neural} approximates the solution operator of linear symmetric PDEs by learning a low-rank decomposition of the Green's function $G(x, y)$ on a point cloud. Two coordinate networks predict left- and right-vector bases at each point, and the solution at a query is recovered by integrating the learned Green's function against the source field. The architecture is designed for forward problems whose governing operator is linear, self-adjoint, and admits a Green's function representation. It is not applicable to inverse problems or nonlinear forward problems. So, we apply NGF to four Poisson problems on irregular geometries (fitting domain, gear, nut, screws/bolts) using the default network and training configuration (trained with 3 data-parallel processes, matching the configuration of the released checkpoints) from the original repository\footnote{\url{https://github.com/KAIST-Visual-AI-Group/NGF.git}} without modification.

\section{Additional Results}
\subsection{Noise Robustness} \label{app:noise_rob}

We adopt a relative-Gaussian noise model scaled by the per-sample $L^\infty$ norm of the measurement field. Let $x$ denote a clean measurement vector; the corrupted measurement is:
\begin{equation}
\label{eq:noise_model}
    \tilde{x} = x + \varepsilon,
    \qquad
    \varepsilon \sim \mathcal{N}\!\left(0,\; \bigl(\eta\,\|x\|_\infty\bigr)^2 I\right).
\end{equation}
for noise levels $\eta \in \{0\%, 1\%, 2\%, 5\%, 10\%\}$

The three benchmarks differ in what the measurement vector $x$ is and at what granularity the $L^\infty$ norm is computed.
\begin{itemize}
    \item \textbf{Airfoil.} The measurement is the partial Mach field observed at a fraction of mesh points. A single $\|x\|_\infty$ is computed per sample over all observed points.
    \item \textbf{EIT.} The measurement consists of $L = 20$ paired Dirichlet--Neumann boundary patterns evaluated at $M=272$ sensors. 
    \item \textbf{RTE.} The measurement is the difference Albedo matrix $\Delta\Lambda \in \mathbb{R}^{N_{rec} \times N_{src}}$. Because receivers have very different outgoing flux magnitudes, we use a per-receiver-row $L^\infty$ scale: $\varepsilon_{r,s} \sim \mathcal{N}(0, (\eta\,\|\Delta\Lambda_{r,:}\|_\infty)^2)$.
\end{itemize}

For each benchmark and noise level, we evaluate the trained model under 10 independent noise realizations and report the mean and standard deviation of the median relative $L^2$ error across these seeds; results are summarized in Table~\ref{tab:noise_robustness}. Across all three benchmarks, GAIA degrades gracefully under measurement noise. Seed-to-seed variation remains low relative to the mean at every noise level, showing that this robustness is a stable property of the trained model.

\begin{table}[h]
    \centering
    \caption{Median relative $L^2$ error (\%) under increasing noise levels $\eta$ on the Airfoil, EIT, and RTE inverse benchmarks. Values are reported as mean $\pm$ std over 10 independent noise realizations.}
    \label{tab:noise_robustness}
    \begin{tabular}{lccccc}
        \toprule
        & \multicolumn{5}{c}{Noise level $\eta$} \\
        \cmidrule(lr){2-6}
        Benchmark & $0\%$ & $1\%$ & $2\%$ & $5\%$ & $10\%$ \\
        \midrule
        Airfoil & $0.49 \pm 0.00$ & $0.58 \pm 0.03$ & $1.12 \pm 0.06$ & $3.87 \pm 0.14$ & $6.51 \pm 0.09$ \\
        EIT     & $0.70 \pm 0.00$ & $0.76 \pm 0.01$ & $0.91 \pm 0.01$ & $1.53 \pm 0.03$ & $2.79 \pm 0.04$ \\
        RTE     & $1.42 \pm 0.00$ & $1.53 \pm 0.01$ & $1.83 \pm 0.01$ & $3.57 \pm 0.03$ & $10.16 \pm 0.09$ \\
        \bottomrule
    \end{tabular}
\end{table}

\subsection{Discretization invariance} 
\label{app:di}
\begin{figure}[!htb]
    \centering
\includegraphics[width=0.4\linewidth]{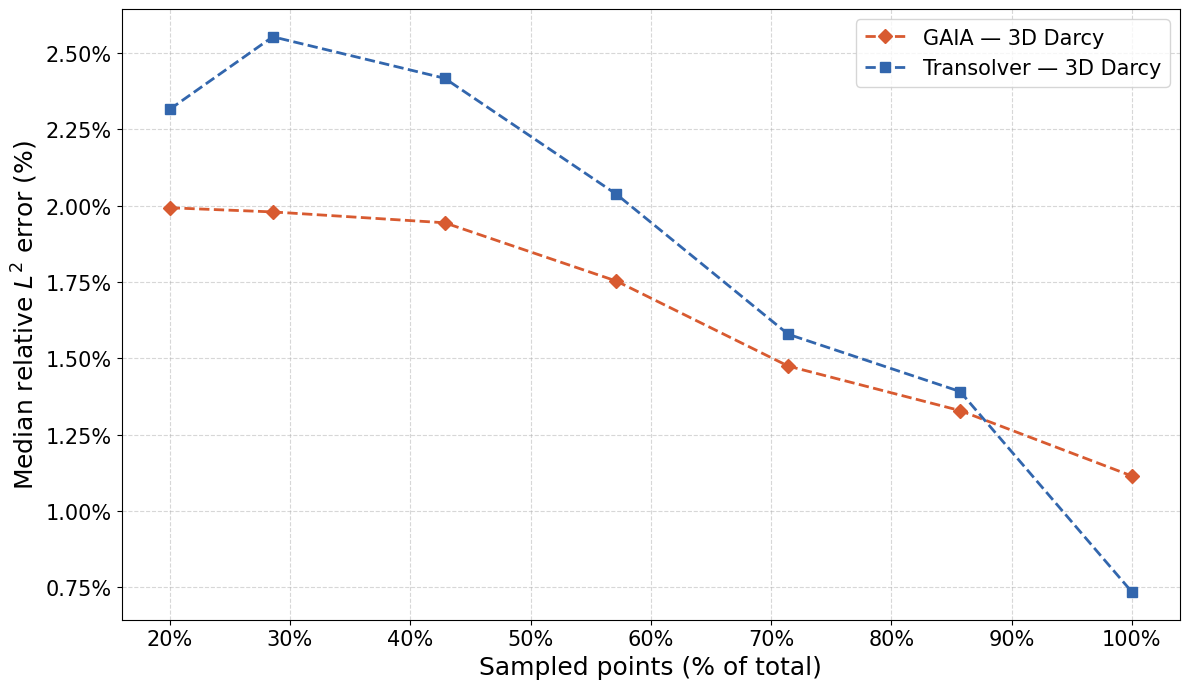} \includegraphics[width=0.4\textwidth]{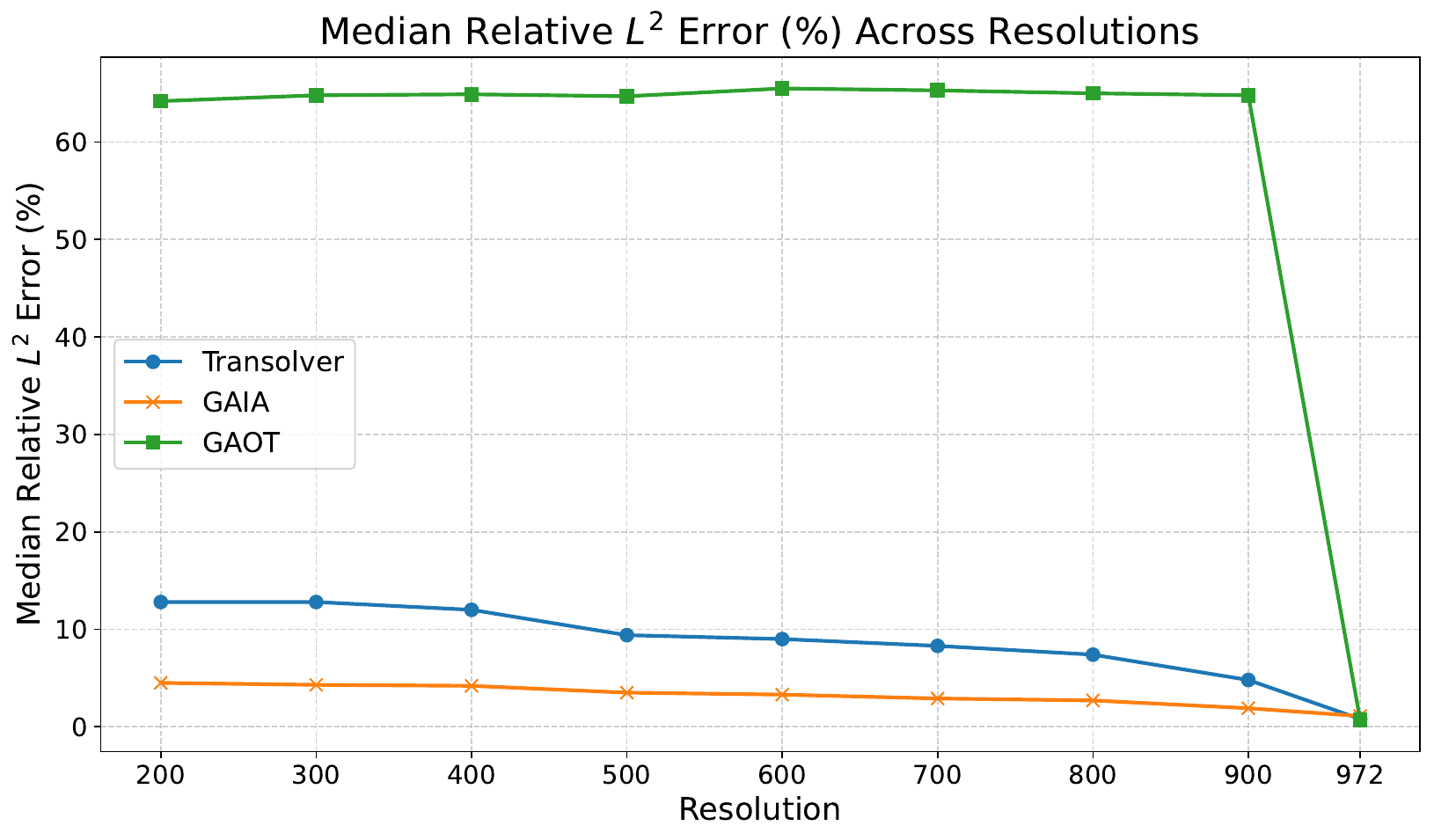}
    \caption{Discretization invariance on 3D Darcy (trained on 7000 points) and on Elasticity (trained at 972 points). Although Transolver and GAOT achieve lower error at the training resolution (100\%), GAIA degrades more gradually under subsampling and achieves lower error at coarser resolutions.}
    \label{fig:di_darcy}
\end{figure}

We conduct a discretization invariance study for Elasticity and 3D Darcy benchmarks (see Figure \ref{fig:di_darcy}). While Transolver's degradation is less severe than for the Elasticity benchmark, GAIA still outperforms Transolver at non-training resolutions. GAOT's performance also suffers under varying resolutions. We used GAOT's official implementation unmodified with their given Elasticity configuration file.  All models were evaluated at the same subsampled coordinates used for GAIA and Transolver, with no per-resolution retraining or tuning for any method. 

\subsection{Data augmentation for discretization invariance} 
\label{ssec:da}

\begin{figure}[!htb]
    \centering
    \includegraphics[width=0.8\textwidth]{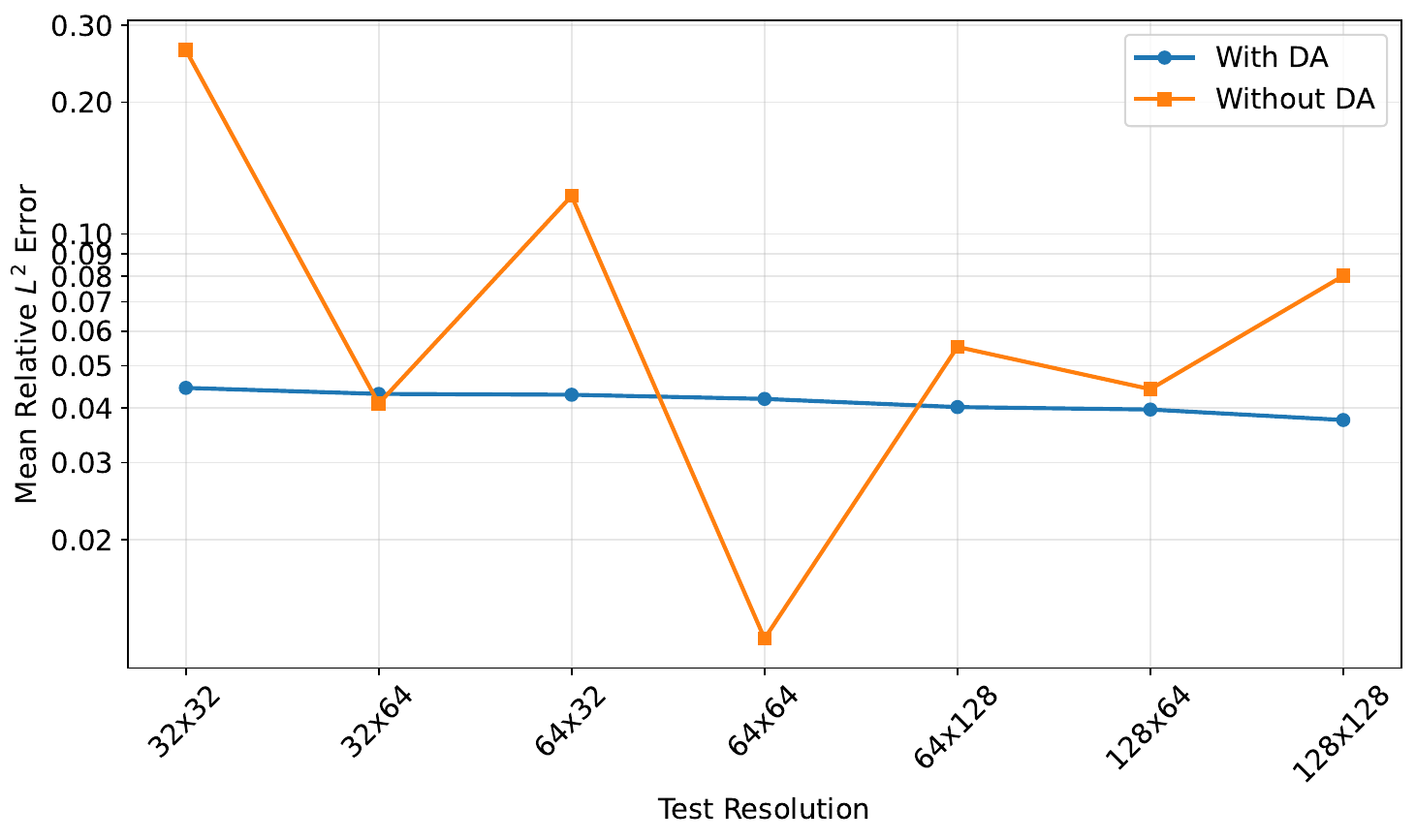}
    \caption{Discretization invariance on Poisson-Gauss. 
    Multi-resolution augmentation (blue) provides one-shot generalization across resolutions compared to single-resolution 
    training (orange).}
    \label{fig:discretization}
\end{figure}
Discretization invariance can be 
further improved by training on multiple resolutions 
simultaneously. We demonstrate this on the Poisson-Gauss benchmark, comparing models trained with and without multi-resolution data augmentation.
The data augmentation used in IAE-Net is implemented here via a stride-based multi-resolution training scheme. The original data is generated on a $128 \times 128$ grid and subsampled to $64 \times 64$ for standard training. For augmented training, the $64 \times 64$ data is upsampled back to the canonical $128 \times 128$ resolution using bicubic interpolation. Multi-resolution views are then generated by striding this grid. Each training iteration simultaneously optimizes over multiple resolutions $\mathcal{R} = \{32 \times 32, 32 \times 64, 64 \times 32, 64 \times 64, 64 \times 128, 128 \times 64, 128 \times 128\}$, with the total loss:
\begin{equation}
    \mathcal{L}_{\text{total}} = \sum_{T \in \mathcal{R}} \lambda_T \cdot \mathcal{L}(\hat{u}_T, u_T),
\end{equation}
where $\lambda_T$ are per-resolution weights (set to 1.0 in this experiment). As shown in Figure~\ref{fig:discretization}, multi-resolution training yields stable performance across all tested resolutions. We note that for unstructured point clouds, generating equivalent 
multi-resolution training views would require scattered data 
interpolation to obtain field values at new point locations, which 
is less straightforward than grid striding and may introduce 
interpolation artifacts.



\subsection{Efficiency analysis} 
\label{ssec:efficiency}

Table~\ref{tab:timing_analysis} reports computational costs on the EIT benchmark ($N=2000$ mesh nodes) measured on a single NVIDIA A6000 GPU in FP32. Latency is the mean wall-clock time over 100 runs (batch size 1) after 20 warm-up iterations with CUDA synchronization; throughput is measured at batch size 16. GAIA achieves competitive latency and memory while maintaining the highest throughput among geometry-adaptive methods. 

To assess the computational efficiency of GAIA relative to baseline methods, we conduct a comprehensive timing analysis on the EIT benchmark with $N=2000$ mesh nodes. All experiments are performed on a single NVIDIA A6000 GPU using FP32 precision.

\textbf{Metrics.} We evaluate the following computational metrics:

\begin{itemize}
    \item \textbf{Parameters:} Total number of learnable weights in the model, reported in millions (M).
    
    \item \textbf{Latency:} Wall-clock time (ms) for a single inference with batch size 1. We report the mean over 100 timed runs following 20 warm-up iterations to ensure stable GPU clock speeds and memory allocation. CUDA synchronization is enforced before each timing measurement to account for asynchronous kernel execution.
    
    \item \textbf{Peak Memory:} Maximum GPU memory allocated during a forward pass with batch size 1, measured after resetting CUDA memory statistics. 
    
    \item \textbf{Throughput:} Maximum number of samples processed per second with batch size 16, measured by timing batched inference over 50 iterations. 
\end{itemize}

\subsection{Comparison with Finite Element Methods}
\label{app:fem_comparison}

We compare the computational cost of GAIA with conventional finite
element methods (FEM) on the 3D Mechanical Components Benchmark (MCB).
While FEM provides higher accuracy, GAIA targets amortized multi-query
settings, such as engineering design, optimization, and inverse
problems, where solutions are required across many geometries or
boundary conditions. In these settings, mesh generation and numerical
solves can contribute substantially to the computational cost.
GAIA operates directly on point clouds without requiring volumetric
mesh generation at inference.

\begin{table}[h]
    \centering
    \caption{Computational cost comparison on MCB. Mesh generation
    times are reported for fTetWild~\cite{yoo2025neural},
    while FEM solve times are measured using FEniCS. GAIA inference
    uses point-cloud inputs and does not require mesh generation.
    Times are reported in seconds.}
    \label{tab:fem_comparison}
    \small
    \setlength{\tabcolsep}{7pt}
    \begin{tabular}{lcccc}
        \toprule
        Category & Mesh gen. & FEM solve & FEM total & GAIA \\
        \midrule
        Fitting       & 18.04 & 0.72 & 18.76 & 0.021 \\
        Gear          & 45.80 & 0.71 & 46.51 & 0.014 \\
        Nut           & 11.60 & 1.14 & 12.74 & 0.025 \\
        Screws/bolts  & 12.58 & 0.55 & 13.13 & 0.019 \\
        \bottomrule
    \end{tabular}
\end{table}

FEM solve times were measured using FEniCS 2019.1.0
(dolfin, PETSc 3.22.2) with the default solver on an
AMD EPYC 9475F CPU. GAIA inference times represent
batch-size-one latency on a single NVIDIA A6000 GPU.

As shown in Table~\ref{tab:fem_comparison}, mesh generation
accounts for most of the conventional FEM pipeline's
computational cost on these geometries. By avoiding this
step and performing single-pass inference, GAIA substantially
reduces per-instance computational cost. This advantage
is particularly relevant when evaluating many previously
unseen geometries, where new meshes would otherwise be
required.

\subsection{Scalability on Large 3D Point Clouds}
\label{app:scaling}

We evaluate the scalability of GAIA on the 3D Darcy benchmark
as the number of input points increases. All experiments are
performed on a single NVIDIA A6000 GPU (48 GB) in FP32.

\paragraph{Training scalability.}
Table~\ref{tab:training_scaling} compares the training-step
latency and peak GPU memory usage of GAIA and Transolver.
Each training step includes the forward pass, backward pass,
and optimizer update. Both methods exhibit approximately
linear scaling in point-cloud size, with memory requirements
approaching the GPU capacity at 400K points.

\begin{table}[h]
\centering
\caption{Training scalability on 3D Darcy. Latency is
reported per training step, including forward and backward
passes and optimizer updates. Memory is reported in MB.}
\label{tab:training_scaling}
\small
\begin{tabular}{rcccc}
\toprule
& \multicolumn{2}{c}{GAIA}
& \multicolumn{2}{c}{Transolver} \\
\cmidrule(lr){2-3}\cmidrule(lr){4-5}
Points & Time (ms) & Mem. (MB) & Time (ms) & Mem. (MB) \\
\midrule
7K   & 57.8   & 930    & 30.0  & 887 \\
50K  & 137.3  & 6,140  & 116.0 & 5,832 \\
100K & 262.2  & 12,220 & 220.2 & 11,555 \\
250K & 681.9  & 30,444 & 533.6 & 28,590 \\
400K & 1082.0 & 48,726 & 848.6 & 45,681 \\
450K & OOM    & OOM    & OOM   & OOM \\
\bottomrule
\end{tabular}
\end{table}

To assess predictive accuracy at higher resolutions, we
additionally train GAIA on 100K-point clouds using 1024
training samples, 256 test samples, and 100 epochs, without
additional hyperparameter tuning. The resulting model achieves
approximately 2.6\% median relative $L^2$ error.

\paragraph{Inference scalability.}
Table~\ref{tab:inference_scaling} reports the inference
latency and GPU memory consumption of GAIA at increasing
point-cloud resolutions.

\begin{table}[H]
\centering
\caption{GAIA inference scalability on 3D Darcy.
Batch-size-one inference on a single NVIDIA A6000 GPU
in FP32.}
\label{tab:inference_scaling}
\small
\begin{tabular}{rcc}
\toprule
Points & Latency (ms) & Memory (MB) \\
\midrule
7K   & 14.18   & 102 \\
50K  & 42.59   & 593 \\
100K & 91.57   & 1,165 \\
250K & 247.00  & 2,879 \\
400K & 397.04  & 4,595 \\
500K & 498.32  & 5,733 \\
1M   & 1002.37 & 11,449 \\
\bottomrule
\end{tabular}
\end{table}

Inference requires substantially less memory than training,
as gradients and optimizer states are not stored. GAIA
supports inference on point clouds containing up to one
million points on a single GPU, requiring approximately
1 second and 11.4 GB of memory. 

\subsection{Training Randomness} \label{app:randomness}
To assess sensitivity to training randomness (weight initialization, data shuffling), we train GAIA and all baselines with different random seeds. Table~\ref{tab:seed_stability} reports the resulting statistics. The small standard deviation indicates stable performance across runs.


\begin{table}[!htb]
    \centering
    \small
    \setlength{\tabcolsep}{3pt}
    \begin{tabular*}{\linewidth}{
        @{\extracolsep{\fill}}lccccc@{}}
        \toprule
        Dataset & GINO & CORAL & LNO & GAOT & GAIA \\
        \midrule
        EIT
        & $26.42 \pm 2.63$
        & $1.65 \pm 0.03$
        & $0.91 \pm 0.04$
        & $8.02 \pm 0.38$
        & $\mathbf{0.71 \pm 0.04}$ \\

        OT
        & $1.90 \pm 0.15$
        & $3.06 \pm 0.16$
        & $1.76 \pm 0.06$
        & $5.79 \pm 0.64$
        & $\mathbf{1.45 \pm 0.04}$ \\

        Airfoil
        & $4.79 \pm 0.69$
        & $1.59 \pm 0.06$
        & $2.49 \pm 0.30$
        & $9.53 \pm 0.34$
        & $\mathbf{0.58 \pm 0.03}$ \\
        \bottomrule
    \end{tabular*}
    \vspace{3mm}
    \caption{Sensitivity to training randomness on inverse
    benchmarks. Mean $\pm$ standard deviation of the test-set
    median relative $L^2$ error (\%) across 5 independent
    training runs.}
    \label{tab:seed_stability}
\end{table}

\begin{table}[H]
    \centering
    \small
    \setlength{\tabcolsep}{3pt}

    \begin{tabular*}{\linewidth}{@{\extracolsep{\fill}}lccccc@{}}
        \toprule
        Dataset & GINO & CORAL & Transolver & GAOT & GAIA \\
        \midrule
        Poisson-Gauss
& $1.210 \pm 0.045$
& $4.360 \pm 0.110$
& $1.220 \pm 0.310$
& $1.372 \pm 0.194$
& $\mathbf{0.770 \pm 0.040}$ \\

Elasticity
& $1.930 \pm 0.040$
& $1.980 \pm 0.048$
& $1.000 \pm 0.130$
& $\mathbf{0.994 \pm 0.004}$
& $1.393 \pm 0.053$ \\

3D Darcy
& $20.510 \pm 0.670$
& $21.580 \pm 0.220$
& $\mathbf{0.789 \pm 0.049}$
& $36.266 \pm 2.020$
& $1.240 \pm 0.170$ \\
        \bottomrule
    \end{tabular*}
\vspace{3mm}
    \caption{Sensitivity to training randomness on forward
    benchmarks. Mean $\pm$ standard deviation of the test-set
    median relative $L^2$ error (\%) across 5 independent
    training runs.}
    \label{tab:forward_seeds}
\end{table}

\begin{table}[htbp]
    \centering
    \small
    \setlength{\tabcolsep}{5pt}

    \begin{tabular*}{\linewidth}{
        @{\extracolsep{\fill}}lcccc@{}}
        \toprule
        Shape & CORAL & NGF & LNO & GAIA \\
        \midrule
        Fitting
        & $18.81 \pm 0.22$
        & $11.25 \pm 0.59$
        & $14.88 \pm 0.58$
        & $\mathbf{10.30 \pm 0.49}$ \\

        Gear
        & $13.81 \pm 0.65$
        & $5.33 \pm 0.51$
        & $5.91 \pm 0.20$
        & $\mathbf{3.63 \pm 0.30}$ \\

        Nut
        & $20.69 \pm 1.82$
        & $5.34 \pm 0.13$
        & $6.95 \pm 0.36$
        & $\mathbf{4.95 \pm 0.25}$ \\

        Screws/Bolts
        & $20.47 \pm 0.38$
        & $6.31 \pm 0.30$
        & $5.22 \pm 0.28$
        & $\mathbf{3.45 \pm 0.36}$ \\
        \bottomrule
    \end{tabular*}
\vspace{3mm}
    \caption{Sensitivity to training randomness on MCB.
    Mean $\pm$ standard deviation of the test-set median
    relative $L^2$ error (\%) across 3 independent training runs.}
    \label{tab:mcb_seeds}
\end{table}

\subsection{Additional Baselines on MCB}
\label{app:mcb_baselines}

We additionally evaluate GINO and GAOT on the 3D Poisson
boundary value problem using the Mechanical Components
Benchmark (MCB). Table~\ref{tab:mcb_additional} reports
their performance alongside GAIA. Both methods are non-competitive in all four
shape categories.

\begin{table}[h]
    \centering
    \caption{Additional baseline comparisons on MCB.
    Median relative $L^2$ error (\%).}
    \label{tab:mcb_additional}
    \small
    \begin{tabular}{lcccc}
        \toprule
        Model & Fitting & Gear & Nut & Screws/Bolts \\
        \midrule
        GINO & 45.08 & 34.89 & 26.19 & 52.83 \\
        GAOT & 52.96 & 47.72 & 51.98 & 73.62 \\
        GAIA & \textbf{10.30} & \textbf{3.63} &
               \textbf{4.95} & \textbf{3.45} \\
        \bottomrule
    \end{tabular}
\end{table}

\section{Ablations}
\label{app:ablate}

We examine the sensitivity of the model to the tokenization hyperparameters, and provide details on the choice of conditioning mechanism, and the contribution of each token pathway.

First, we conduct ablations on every design module across benchmarks. We also conduct studies on the EIT benchmark to investigate the sensitivity of GAIA to the token dimension and the total number of tokens of the tokenization module. We additionally ablate the geometry conditioning mechanism itself.

\subsection{Architectural Component Ablations}
\label{app:architecture_ablation}

We evaluate the contribution of individual architectural components
across four benchmarks: Elasticity, EIT, Airfoil, and Optical Tomography
(RTE). Each ablation modifies one component of GAIA while keeping the
remaining architecture and training protocol unchanged.

\begin{table}[h]
    \centering
    \caption{Architectural component ablations across four benchmarks.
    Median relative $L^2$ error (\%).
    Each variant modifies one component of the full model. Elasticity has no boundary$\rightarrow$interior transfer, so O2D does not apply.}
    \label{tab:architecture_ablation}
    \small
    \setlength{\tabcolsep}{5pt}
    \begin{tabular}{lcccc}
        \toprule
        Variant & Elasticity & EIT & Airfoil & RTE \\
        \midrule
        GAIA (full) & \textbf{1.39} & \textbf{0.71} &
        \textbf{0.58} & 1.45 \\
        Refined tokens $\rightarrow$ static tokens
            & 1.52 & 0.73 & 0.92 & 1.49 \\
        PointNet $\rightarrow$ linear pooling
            & 1.49 & 0.85 & 0.82 & \textbf{1.32} \\
        Dense skips $\rightarrow$ ResNet skips
            & 1.36 & 0.96 & 0.75 & 1.50 \\
        Dense skips $\rightarrow$ no skips
            & 1.83 & 1.41 & 4.13 & 3.07 \\
        Fourier features $\rightarrow$ raw coordinates
            & 1.91 & 1.03 & 1.72 & 1.43 \\
        O2D decoder $\rightarrow$ direct boundary map
            & -- & 1.15 & 0.73 & \textbf{1.31} \\
        \bottomrule
    \end{tabular}
\end{table}

Table~\ref{tab:architecture_ablation} shows that the full model is the best on almost every benchmark; no reduced variant is uniformly competitive. Skip connections contribute significantly, removing them consistently degrades performance. Static tokens cost more on diverse geometry families (Elasticity, Airfoil, RTE), indicating token refinement matters. Direct boundary-to-interior (removing the O2D latent set) is often less accurate.

Intuitively, each module addresses a specific property of the operator being learned. Cross-attention kernel conditioning supports the integral kernel solution operator’s spatially adaptive dependence on the domain. Dual-pathway tokenizer provides disentangled physics and geometry representations. Fourier positional features help capture high-frequency spatial dependence, which coordinate MLPs represent poorly under spectral bias. Dense skips preserve multi-scale features across the block stack, since the encode-process-decode bottleneck in each block may discard fine detail that later blocks need. Token refinement between blocks lets the conditioning signal evolve with the features. The O2D decoder transfers
sparse boundary measurements to the interior through a fixed
set of $K$ latent modes, reducing the cost of a direct
$O(MN)$ boundary-to-interior integral transform to
$O(K(M+N))$, where $M$ and $N$ denote the numbers of boundary
and interior points, respectively.  The boundary encoder is permutation-invariant and cardinality-agnostic so that the shape representation does not depend on the ordering and resolution of boundary samples.

\subsection{Token Dimension}
Table \ref{tab:ablation_token_dim} reports the median relative
$L^2$ error as the token dimension varies from 4 to 64, with the number of tokens fixed at 8. Performance is best at 4 and degrades gradually as the dimension increases, suggesting that a compact token representation is sufficient to capture the relevant geometric information for this benchmark. Larger token dimensions may introduce unnecessary capacity that hinders optimization without providing additional representational benefit.
\begin{table}[h]
\centering
\begin{tabular}{lrrrrr}
\toprule
Token Dim & 4 & 8 & 16 & 32 & 64 \\
\midrule
L2 Median & \textbf{0.00698} & 0.00712 & 0.00737 & 0.00804 & 0.00832 \\
\bottomrule
\end{tabular}
\caption{Effect of token dimension with number of tokens fixed to 8.}
\label{tab:ablation_token_dim}
\end{table}

\subsection{Number of Geometric Tokens}
Table \ref{tab:ablation_num_tokens} examines the effect of varying the token count from 4 to 64 with the token dimension fixed at 8. Performance improves sharply from 4 to 8 tokens, after which it plateaus. This indicates that for simple shapes, modest number of tokens is sufficient to encode the geometric structure of the domain, and that additional tokens do not meaningfully improve accuracy.

\begin{table}[h]
\centering
\begin{tabular}{lrrrr}
\toprule
Num Tokens & 4 & 8 & 16 & 64 \\
\midrule
L2 Median & 0.00851 & \textbf{0.00712} & 0.00727 & 0.00760 \\
\bottomrule
\end{tabular}
\caption{Effect of number of tokens with token dimension fixed to 8.}
\label{tab:ablation_num_tokens}
\end{table}

\subsection{Number of Slice Tokens}
The slice tokenizer aggregates the input
sensor data into a fixed set of physics-informed tokens via softmax-weighted
clustering. Table~\ref{tab:ablation_num_slice_tokens} examines the effect of
varying the slice-token count from 32 to 96, with all
other hyperparameters held fixed. Performance is best at 64 slice tokens and
degrades only mildly outside this setting, indicating that a moderate number
of slices is sufficient to summarize the boundary measurements without
introducing redundancy. 

\begin{table}[h]
\centering
\begin{tabular}{lrrrr}
\toprule
Num Slice Tokens & 32 & 48 & 64 & 96 \\
\midrule
L2 Median & 0.00797 & 0.00873 & \textbf{0.00712} & 0.00735 \\
\bottomrule
\end{tabular}
\caption{Effect of the number of slice tokens on the EIT benchmark, with all
other hyperparameters fixed.}
\label{tab:ablation_num_slice_tokens}
\end{table}

\subsection{Number of Blocks}
Table~\ref{tab:ablation_num_blocks} examines the effect of
varying the number of blocks from 3 to 7 on the EIT benchmark, with all
other hyperparameters held fixed. Performance is best at 5 blocks and
remains close at neighbouring depths, indicating that the model is robust to
this choice.

\begin{table}[h]
\centering
\begin{tabular}{lrrrrr}
\toprule
Num Blocks & 3 & 4 & 5 & 6 & 7 \\
\midrule
L2 Median & 0.00762 & 0.00789 & \textbf{0.00712} & 0.00739 & 0.008406\\
\bottomrule
\end{tabular}
\caption{Effect of number of blocks with all other hyperparameters fixed.}
\label{tab:ablation_num_blocks}
\end{table}

\subsection{Geometry Conditioning Ablation Details}
Table \ref{tab:ablations} compares three alternatives for using geometry information to condition the integral kernel on a forward problem (Elasticity) and an inverse problem (Airfoil). In \textit{concatenation}, the geometry tokens are mean-pooled into a single vector, projected, and concatenated onto the per-point feature before entering the kernel MLP, thus providing global context but no spatial selectivity. In \textit{FiLM} conditioning, the mean-pooled token vector is projected to produce per-channel scale and shift parameters that affinely modulate the combined feature representation. The proposed {cross-attention} mechanism instead lets each spatial query point attend independently to the full set of geometry tokens via multi-head attention. Cross-attention consistently achieves the lowest error on both benchmarks, implying that spatially adaptive conditioning is important for capturing local geometric features.

\subsection{Tokenizer Ablation Details} 
 The IAE refinement stage stacks several
adaptive blocks, each applying integral transforms followed by token-updated
cross-attention. Table \ref{tab:ablations} evaluates the contribution of each token pathway by removing either the boundary feature tokens (produced by the PointNet-style encoder from $X_{\text{bnd}}$) or the slice tokens (produced by the soft-clustering encoder from the interior mesh and field values). Both ablations increase error relative to the full model on the Elasticity and Airfoil benchmarks, implying that both tokenization pathways contribute information: boundary tokens capture the global domain shape, while slice tokens encode the spatially varying physics of the input field.




\end{document}